\documentclass[runningheads]{llncs}

\usepackage{eccv}

\usepackage{eccvabbrv}

\usepackage{graphicx}
\usepackage{booktabs}

\usepackage[accsupp]{axessibility}  

\usepackage{hyperref}

\usepackage{orcidlink}
\usepackage{graphicx}
\usepackage{multirow}
\usepackage{tikz}
\usepackage{comment}
\usepackage{amsmath,amssymb} 
\usepackage{color}
\usepackage{wrapfig} 
\usepackage{booktabs}
\usepackage{multirow}
\usepackage{amssymb}
\usepackage{bbding}
\newcommand{\aka}{\textit{a.k.a.}}

\begin{document}

\title{Learning Modality-agnostic Representation for Semantic Segmentation from Any Modalities}
\titlerunning{Learning Modality-agnostic Representation for Semantic Segmentation}

\author{Xu Zheng\inst{1}\orcidlink{0000-0003-4008-8951} \and
Yuanhuiyi Lyu\inst{1}\orcidlink{0009-0004-1450-811X} 
\and
Lin Wang\inst{1,2}\orcidlink{0000-0002-7485-4493}\thanks{Corresponding author}}

\authorrunning{X. Zheng et al.}

\institute{Hong Kong University of Science and Technology, Guangzhou, China \\
\email{zhengxu128@gmail.com, yuanhuiyilv@hkust-gz.edu.cn}
\and
Hong Kong University of Science and Technology, Hong Kong, China \\
\email{linwang@ust.hk}\\
\url{https://vlislab22.github.io/Any2Seg/} 
}

\maketitle

\begin{abstract}
Image modality is not perfect as it often fails in certain conditions, \eg, night and fast motion. This significantly limits the robustness and versatility of existing multi-modal (\ie, Image+X) semantic segmentation methods when confronting modality absence or failure, as often occurred in real-world applications. Inspired by the open-world learning capability of multi-modal vision-language models (MVLMs), we explore a new direction in learning the modality-agnostic representation via knowledge distillation (KD) from MVLMs. Intuitively, we propose \textbf{\textit{Any2Seg}}, a novel framework that can achieve robust segmentation from \textbf{\textit{any}} combination of modalities in \textbf{\textit{any}} visual conditions. Specifically, we first introduce a novel language-guided semantic correlation distillation (\textbf{LSCD}) module to transfer both inter-modal and intra-modal semantic knowledge in the embedding space from MVLMs, \eg, LanguageBind~\cite{zhu2023languagebind}. This enables us to minimize the modality gap and alleviate semantic ambiguity to combine any modalities in any visual conditions. Then, we introduce a modality-agnostic feature fusion (\textbf{MFF}) module that reweights the multi-modal features based on the inter-modal correlation and selects the fine-grained feature. This way, our Any2Seg finally yields an optimal modality-agnostic representation. Extensive experiments on two benchmarks with four modalities demonstrate that Any2Seg achieves the state-of-the-art under the multi-modal setting (\textbf{+3.54} mIoU) and excels in the challenging modality-incomplete setting(\textbf{+19.79} mIoU).  
\end{abstract}

\begin{figure}[h!]
    \centering
    \includegraphics[width=\linewidth]{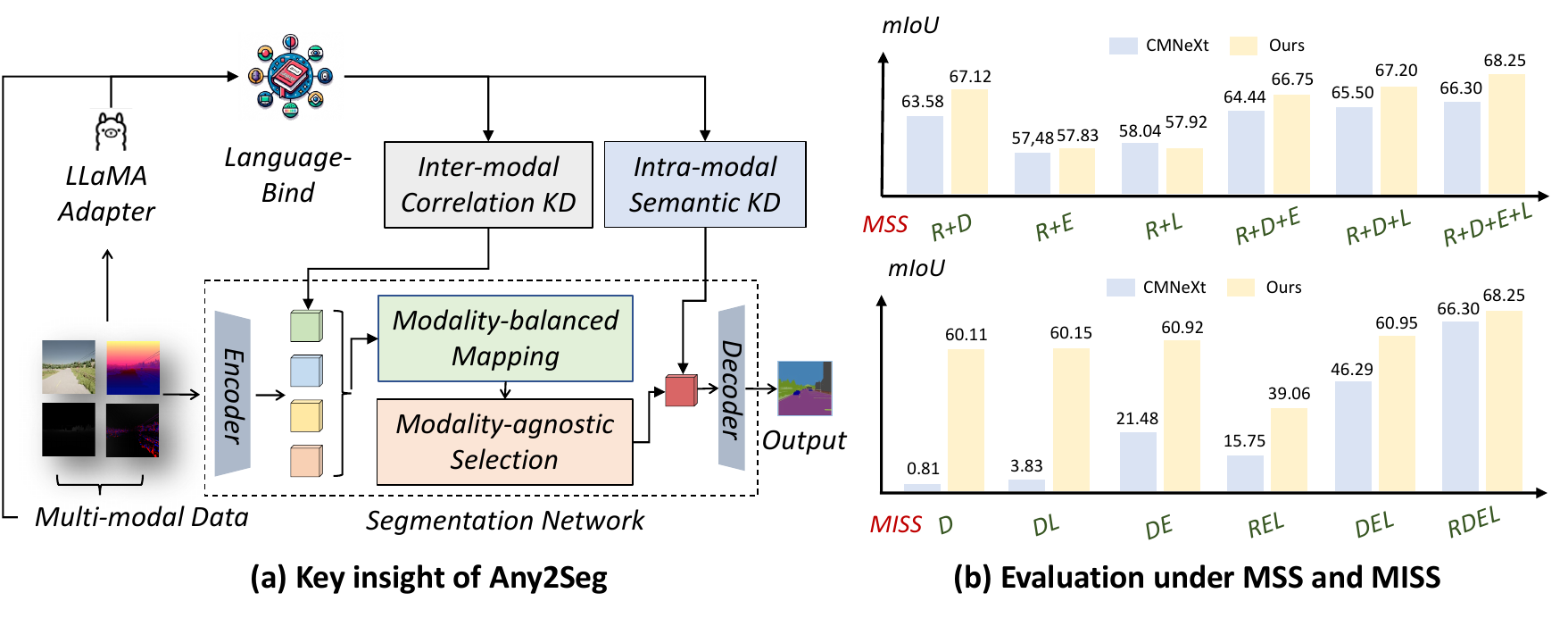}
    \caption{
    (a) Our \textbf{Any2Seg} aims to learn a modality-agnostic representation for robust segmentation from any modalities guided by multi-modal VLMs; (b) Performance comparison between CMNeXt~\cite{zhang2023delivering} and our \textbf{Any2Seg} under the multi-modal semantic segmentation (MSS) and modality-incomplete semantic segmentation (MISS) settings.
    }
    \label{teaser}
\end{figure}

\section{Introduction} 
The integration of multi-modal sensory input is pivotal for enabling intelligent agents to achieve robust and reliable scene understanding, \eg, semantic segmentation~\cite{huang2022multi,wang2023multi,milioto2019rangenet++,zhang2020polarnet,alonso2019ev,jia2023event,sun2019rtfnet,wang2016learning,park2017rdfnet,zheng2023distilling,zheng2024360sfuda++,zheng2024semantics,chen2024frozen,chen2023clip,zheng2022transformer,zhu2023good,zheng2023look,zheng2023both}.
As such, distinct visual modalities have been introduced into multi-modal systems, such as depth~\cite{wang2020learning,zhou2020rgb,wang2020deep,cao2021shapeconv,chen2021spatial}, LiDAR~\cite{zhuang2021perception,yan20222dpass,wang2022multimodal}, and event~\cite{alonso2019ev,zhang2021issafe}, to accompany RGB data to achieve robust scene understanding.
Recently, endeavors have been made in scaling from dual modality to multiple modality fusion which is capable of processing a broader spectrum of input modalities~\cite{zhang2022cmx,zhang2023delivering,zhu2023visual,zheng2023deep,zhou2023clip,zheng2024eventdance}, \aka, multi-modal fusion frameworks. 

However, these methods predominantly rely on image modality, which is not always reliable as it often fails in certain conditions, \eg, extreme lighting. This significantly limits the robustness and versatility of multi-modal (\ie, RGB+X modality) frameworks~\cite{zhang2022cmx,song2022improving, zhang2023delivering, liu2024fourier,cao2023chasing} 
when confronting modality absence and failure, as often occurred in the wild. Humans can automatically combine sensory modalities when understanding the environment. Intuitively, for machines to emulate humans, the multi-modal segmentation model should also be able to dynamically combine various modalities across different environments. This is particularly beneficial for overcoming modality absence and failure challenges, \aka, \textbf{system-level} and \textbf{sensor-level modality-incomplete} conditions~\cite{liu2024fourier}.

The recent multi-modal vision-language models (MVLMs) have gained significant attention for their strong open-world learning capability and superiority in alleviating cross-modal heterogeneity by binding multiple visual modalities to a central modality (\eg, language) in the representation space~\cite{han2023imagebind, zhu2023languagebind,guo2023point,zhang2023llama, zhang2022pointclip,zhou2024exact,lyu2024unibind,lyu2024omnibind}. 
Inspired by their success, we explore a \textbf{new} direction by learning modality-agnostic representation to tackle aforementioned challenges under the guidance of MVLMs. 
Intuitively, we present \textbf{\textit{Any2Seg}}, a novel multi-modal segmentation framework that can achieve robust semantic segmentation from any combination of modalities in any visual conditions, as depicted in Fig.~\ref{teaser} (a). 

To make this possible, we first propose a language-guided semantic correlation distillation (\textbf{LSCD}) module \textbf{to minimize the modality gap and further alleviate semantic ambiguity} to achieve the capability of combining any modalities in any visual conditions (Sec.~\ref{LCSD}). 
Specifically, to minimize the modality gap, LSCD first distills the inter-modal correlation with text descriptions of the targeted scene generated by LLaMA-adapter~\cite{zhang2023llama}. Then, to alleviate semantic ambiguity, LSCD extracts the intra-modal semantic knowledge from LanguageBind and transfers it to the segmentation model. 


With the transferred inter- and intra-modal semantic knowledge, we then introduce a novel Modality-agnostic Feature Fusion (\textbf{MFF}) module to \textbf{learn an optimal modality-agnostic representation} from the multi-modal features (Sec.~\ref{sec:MFFM}).
Concretely, the inter-modal semantic correlation from the LSCD is utilized to reweight the multi-modal features. Next, we select the fine-grained feature across multi-modal features for better fusion from any combination of modalities. Finally, MFF aggregates the reweighted features and the fine-grained feature to learn the optimal modality-agnostic representation.

We conduct extensive experiments across dual, triple, and quadruple modality combinations on DLIVER~\cite{zhang2023delivering} and MCubeS~\cite{liang2022multimodal} datasets with two experimental settings: 1) \textit{multi-modal segmentation} with all modalities as inputs, and 2) \textit{modality-incomplete segmentation} which accepts any combination of modalities in any visual conditions during the inference. 
Experimental results demonstrate that Any2Seg significantly surpasses prior SoTA methods~\cite{zhang2023delivering, broedermann2022hrfuser, zhang2023cmx, liu2024fourier}, under the multi-modal segmentation setting (See Fig.~\ref{teaser} (b)). 
When extending to challenging modality-incomplete settings that accommodate inputs from four modalities, our Any2Seg significantly outperforms existing approaches, achieving a remarkable enhancement of \textbf{+19.79} mIoU.

In summary, the main contributions of our paper are as follows: 
\textbf{(I)} We make the \textbf{\textit{first}} attempt to explore the potential of MVLMs and propose Any2Seg that can achieve robust segmentation from any combination of modalities in any visual conditions by distilling knowledge from MVLMs;
\textbf{(II)} We propose a novel LSCD module to achieve the capability of combining any modalities in any visual conditions;
\textbf{(III)} We introduce a novel MFF module to learn the optimal modality-agnostic representation;
\textbf{(IV)} We conduct extensive experiments on two multi-modal benchmarks with four modalities in two settings, and the results demonstrate the superiority and robustness of our Any2Seg.

\section{Related Work}
\noindent \textbf{Multi-modal Semantic Segmentation}
Multi-modal semantic segmentation aims at improving the performance by introducing more visual modalities that consists of complementary scene information. Endeavors have been made in introducing distinct visual modalities, such as depth~\cite{wang2020learning,zhou2020rgb,wang2020deep,cao2021shapeconv,chen2021spatial,ying2022uctnet,lee2022spsn,cong2022cir,ji2022dmra,wang2022learning,song2022improving}, LiDAR~\cite{zhuang2021perception,yan20222dpass,wang2022multimodal,li2022deepfusion,borse2023x,zhang2023mx2m,liu2022camliflow,li2023mseg3d}, event~\cite{alonso2019ev,zhang2021issafe}, and thermal data~\cite{shivakumar2020pst900,zhang2021abmdrnet,wu2022complementarity,liao2022cross,zhou2023mmsmcnet,xie2023cross,chen2022modality,pang2023caver,hui2023bridging,zhang2023efficient}, to accompany the RGB data to achieve robust scene understanding. Recently, with the development of novel sensors, such as the event cameras~\cite{zheng2023deep}, various approaches have been proposed to scale from dual modality fusion to multiple modality fusion for achieving robust scene understanding at all day time~\cite{broedermann2022hrfuser,wei2023mmanet,zhang2021abmdrnet,man2023bev,wang2022multimodal,chen2021spatial,zhang2023delivering,zhang2022cmx}. CMNeXt~\cite{zhang2023delivering} is a representative work that achieves multi-modal semantic segmentation with arbitrary-modal complements by taking RGB as fundamentals and other modalities as auxiliary inputs. 

The most significant challenge in multi-modal semantic segmentation is how to alleviate the cross-modal heterogeneity while efficiently extract comprehensive scene information. Existing works always design tailored components to overcome the modality gaps, such as the self-query hub in CMNeXt~\cite{zhang2023delivering}. However, this kind of component only gives marginal performance gain and shows unsatisfactory performance when combining more modalities.
Differently, inspired by the success of MVLMs, \textit{we pioneeringly unleash the potential of MVLMs} and take them as teacher to minimize the modality gap and reduce semantic ambiguity for better multi-modal segmentation.


\noindent \textbf{Multi-modal Vision-language Models (MVLMs).}
The integration of diverse modalities into a unified embedding space for cross-modal learning has emerged as a burgeoning research area. Initial efforts, exemplified by CLIP~\cite{radford2021learning}, have employed contrastive learning paradigms to align image and text pairs, aiming to achieve promising zero-shot generalization performance. Recent advancements have further expanded the scope of learning large-scale vision-language models to encompass multiple modalities, encompassing video~\cite{sun2023learning,cheng2023cico,zheng2023cvt}, point cloud~\cite{zhang2022pointclip,zhu2022pointclip,huang2022clip2point,wang2022multimodal}, thermal~\cite{girdhar2023imagebind,zhang2023cmx}, event~\cite{zhou2023clip,zheng2023deep}, and other modalities.

ImageBind~\cite{girdhar2023imagebind} leverages the binding property of image modality to bind the multi-modal embeddings with the image modality. Subsequent works~\cite{zhang2023meta, su2023pandagpt,han2023imagebind}, such as Point-Bind~\cite{guo2023point}, extend ImageBind~\cite{han2023imagebind} to integrate point cloud data, resulting in remarkable zero-shot 3D capabilities. 
More recently, LanguageBind~\cite{zhu2023languagebind} employs language modality as the binding center across various modalities, aiming to address multi-modal tasks. 
In this paper, we make the \textbf{first} attempt to\textit{ explore the potential of MVLMs and propose Any2Seg that can achieve robust segmentation from any combination of modalities in any visual conditions by distilling knowledge from MVLMs}.

\noindent \textbf{Modality-Incomplete Semantic Segmentation}
Research endeavors have been made in investigating robust frameworks that accommodate complete modality presence during training, despite the potential random absence of modalities during validation ~\cite{liu2024fourier,wang2023multi,maheshwari2024missing,wang2023unibev,reza2023robust,chen2023redundancy,zhao2023multi}.
To address the challenge of missing modalities in multi-modal tasks, ShaSpec \cite{wang2023multi} has been designed to exploit all available inputs by learning shared and specific features, facilitated by a shared encoder alongside individual modality-specific encoders. 
Wang \etal \cite{wang2023learnable} introduced a method to adaptively discern crucial modalities and distill knowledge from them, aiding in the cross-modal compensation for absent modalities. 

More recently, Liu \etal \cite{liu2024fourier} have broadened the scope by establishing the concept of modality-incomplete scene segmentation, addressing both system-level and sensor-level modality deficiencies. They introduced a missing-aware modal switch strategy that mitigates over-reliance on predominant modalities in multi-modal fusion and proactively regulates missing modalities during the training phase. \textit{Different from the ideas to achieve cross-modal compensation and design missing-aware strategies, \textbf{we directly focus on learning modality-agnostic representation}.} Our Any2Seg prioritizes knowledge distillation from MVLMs and fine-grained feature fusion, thereby ensuring robustness and efficiency in both multi-modal and modality-incomplete segmentation tasks.

\section{The Proposed Any2Seg Framework}
\subsection{Overview}
\begin{figure}[t!]
    \centering
    \includegraphics[width=\linewidth]{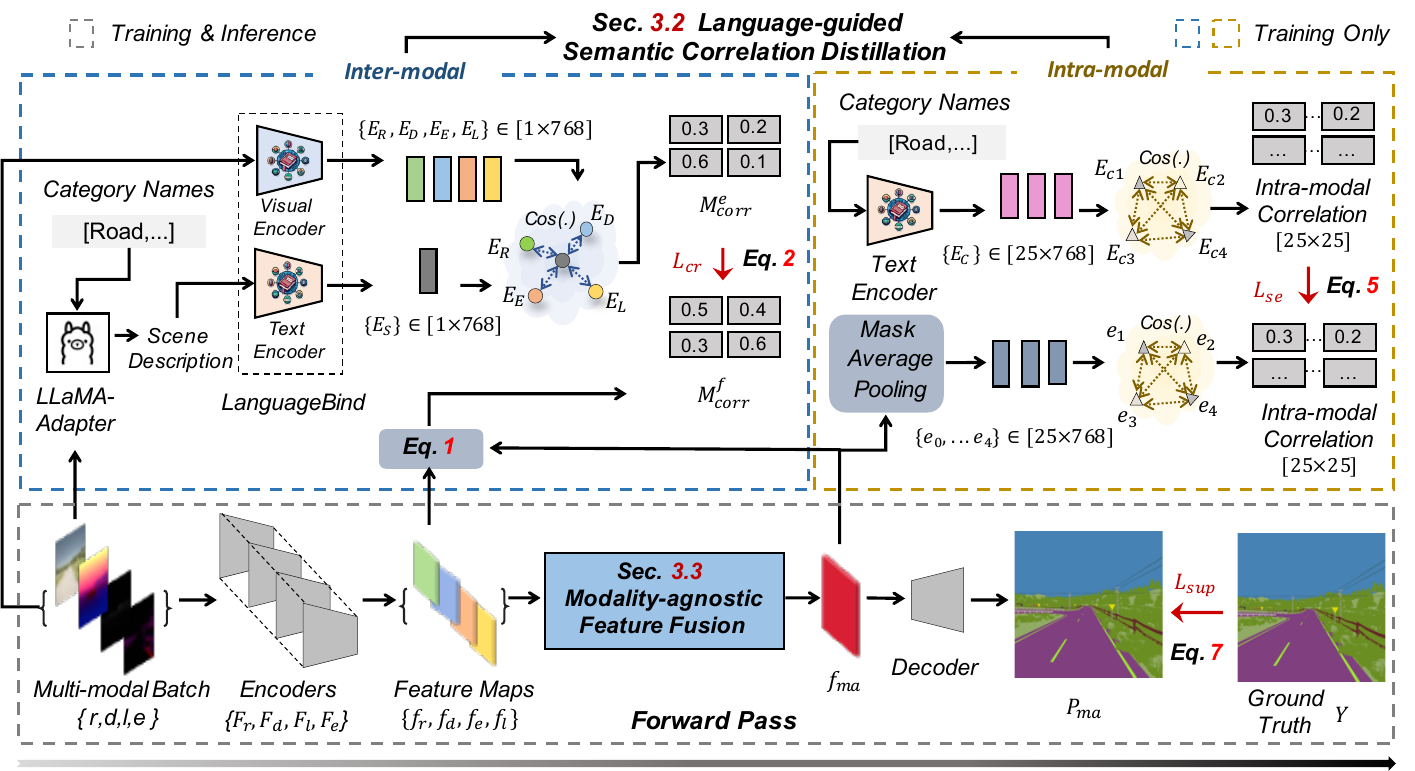}
    \caption{
    The overall framework of \textbf{Any2Seg}, incorporating the LSCD and MFF modules to learn an optimal modality-agnostic representation for robust segmentation.
    }
    \label{overall}
\end{figure}

An overview of our Any2Seg is depicted in Fig.~\ref{overall}. 
It consists of two key modules: the Language-guided Semantic Correlation Distillation (\textbf{LSCD}) (Sec~\ref{LCSD}) and the Modality-agnostic Feature Fusion (\textbf{MFF}) (Sec~\ref{sec:MFFM}) modules. The LSCD module is only for training, and the MFF module is utilized in both training and inference to aggregate multi-modal feature maps \(\{f_r, f_d, f_e, f_l\}\) into modality-agnostic feature \(f_{ma}\), as shown in the forward pass of Fig.~\ref{overall}.

\noindent \textbf{Inputs:} Our Any2Seg processes multi-modal visual data from four modalities, all within the same scene. We consider RGB images \(\textbf{R} \in \mathbb{R}^{h \times w \times 3}\), depth maps \(\textbf{D} \in \mathbb{D}^{h \times w \times C^D}\), LiDAR data \(\textbf{L} \in \mathbb{L}^{h \times w \times C^P}\), and event stack images \(\textbf{E} \in \mathbb{E}^{h \times w \times C^E}\) to illustrate our method, as depicted in Fig.~\ref{overall}. Here, we follow the data processing as ~\cite{zhang2023delivering}, where the channel dimensions $C^D=C^P=C^E=3$. Any2Seg also integrates the corresponding ground truth \(Y\) across \(K\) categories. For each training iteration, Any2Seg takes a mini-batch \(\{r,d,e,l\}\) containing samples from all the input modalities.

\noindent \textbf{MVLM and Backbone:} Within Any2Seg, we integrate the cutting-edge LanguageBind~\cite{zhu2023languagebind}, functioning as a ‘teacher' for knowledge distillation. We incorporate the multi-modal encoders of ~\cite{zhu2023languagebind} to extract corresponding embeddings for all input modalities, as shown in the visual and the text encoders in Fig.~\ref{overall}. 
Concurrently, we employ SegFormer~\cite{xie2021segformer} as the segmentation backbone.

\noindent \textbf{Outputs:} During the inference, Any2Seg utilizes only the forward pass, as depicted in Fig.~\ref{overall}. Given a multi-modal data mini-batch \(\{r,d,e,l\}\), the data is first fed into the encoder of the segmentation backbone. The high-level feature maps \(\{f_r, f_d, f_l, f_e\}\) are then passed to our MFF module for extracting the modality-agnostic feature map \(f_{ma}\). Finally, the segmentation head utilizes \(f_{ma}\) to generate the predictions.
We now describe the two proposed modules in detail.
\subsection{Language-guided Semantic Correlation Distillation (LCSD)}
\label{LCSD}
Building on the recent MVLMs have the strong open-world learning capability and superiority in alleviating cross-modal heterogeneity, we advocate for the knowledge distillation from these cutting-edge MVLM, \ie, LanguageBind~\cite{zhu2023languagebind}. It takes language modality as the central modality to bind all the visual modalities at the high-level representation space.  

\noindent \textbf{Scene Description Generation.} 
To obtain the text description of the targeted scene, we use the LLaMA-Adapter \cite{zhang2023llama} to generate a detailed scene description with visual data and category names.
This model adeptly crafts linguistic representation of scenes, exemplified by \texttt{[`a road that goes along...']}, from visual input batch \(\{r,d,e,l\}\) and their corresponding category names, such as \texttt{[`Road']}. Following this, the category names, along with the generated scene description and the visual input batch \(\{r,d,e,l\}\), are fed into the text and visual encoders of LanguageBind~\cite{zhu2023languagebind}, respectively. This results in a series of high-level feature embeddings, including semantic embeddings \(E_C\) from category names, multiple visual embeddings \(\{E_R,E_D,E_E,E_L\}\), and scene embedding \(E_S\). \footnote{Exact operations can be found in the supplementary material.}

\noindent \textbf{Inter-modal Correlation Distillation: }
To minimize the modality gap, our Any2Seg employs the visual embeddings \(\{E_r, E_d, E_e, E_l\}\) and scene embedding \(E_S\) to compute inter-modal correlation, as illustrated in Fig.~\ref{overall}. As discussed in LanguageBind~\cite{zhu2023languagebind}, linguistic concepts possess an inherent modality-agnostic nature and are less susceptible to style variations in the visual modalities.
Consequently, we utilize the scene embedding \(E_s\) as a central reference for calculating embedding distances relative to \(\{E_r, E_d, E_e, E_l\}\), leading to the formation of the correlation matrix \(M_{corr}^e\) as follows:
\begin{equation}
\setlength{\abovedisplayskip}{3pt}
\setlength{\belowdisplayskip}{3pt}
M_{corr}^e = Concat(Cos(E_r, E_S), Cos(E_d, E_S), Cos(E_e, E_S), Cos(E_l, E_S)).
\end{equation}
where \({Cos}\) denotes the computation of vector similarities utilizing cosine similarity metrics. 
Furthermore, as shown in the \textit{‘Inter-modal Correlation'} of Fig.~\ref{overall}, the obtained modality-agnostic feature \(f_{ma}\) is employed as an anchor point for calculating the correlation with the multi-modal feature maps \(\{f_r, f_d, f_e, f_l\}\):
\begin{equation}
\setlength{\abovedisplayskip}{3pt}
\setlength{\belowdisplayskip}{3pt}
M_{corr}^f = Concat(Cos(f_r, f_{ma}), Cos(f_d, f_{ma}), Cos(f_e, f_{ma}), Cos(f_l, f_{ma})).
\end{equation}
The loss function for mirroring similarity matrices to transfer the inter-modal correlation knowledge:
\begin{equation}
\setlength{\abovedisplayskip}{3pt}
\setlength{\belowdisplayskip}{3pt}
\mathcal{L}_{cr} = KL(M_{corr}^f, M_{corr}^e),
\label{eq:l_cr}
\end{equation}
where \(KL\) denotes the Kullback-Leibler Divergence.
After minimizing modality gaps among visual modalities, we strive to reduce the semantic ambiguity by transferring the intra-modal semantic knowledge.

\noindent \textbf{Intra-modal Semantic Distillation: } We employ semantic embeddings \(E_c\) which are extracted by the text encoder of LanguageBind~\cite{zhu2023languagebind} from category names, such as \texttt{[`Road']}, as supervision signals for segmentation model, as shown in Fig.~\ref{overall}. 
Subsequently, we use the ground truth \(y \in Y\) and the modality-agnostic feature to compute high-level class-wise representation, as shown in the \textit{Class-wise Rep} in  Fig.~\ref{overall}, with Mask Average Pooling (MAP):
\begin{equation}
\setlength{\abovedisplayskip}{3pt}
\setlength{\belowdisplayskip}{3pt}
\{e_0, \ldots, e_K\} = MAP(f_{ma}, y).
\end{equation}
The detailed MAP process is as follows: 1) we first up-sample the modality-agnostic feature $f_{ma}$ to fit the size of ground truth \(y\); 2) we then use \(y\) as class-wise masks to derive per-class features; 3) lastly, we perform average pooling on these per-class features to get the final class-wise semantic representation \(\{e_0, \ldots, e_K\}\).
Building upon this, LSCD initially transfers intra-modal semantic knowledge with implicit relationship transfer. Specifically, we emulate the self-similarity matrix of \(E_c\) and \(\{e_0, \ldots, e_K\}\) to achieve the intra-modal semantic knowledge distillation:
\begin{equation}
\setlength{\abovedisplayskip}{3pt}
\setlength{\belowdisplayskip}{3pt}
\mathcal{L}_{se} = KL({Cos}(\{e_0, \ldots, e_K\}, \{e_0, \ldots, e_K\}^T), {Cos}(E_C, E_C^T)).
\label{eq:l_se}
\end{equation}
The overall knowledge distillation loss of LSCD can be formulated as:
\begin{equation}
\setlength{\abovedisplayskip}{3pt}
\setlength{\belowdisplayskip}{3pt}
\mathcal{L}_{kd} = \mathcal{L}_{cr} + \mathcal{L}_{se}.
\label{eq:kdloss}
\end{equation}

\subsection{Modality-agnostic Feature Fusion (MFF) Module }
\label{sec:MFFM}
\begin{figure}[t!]
    \centering
     \includegraphics[width=\linewidth]{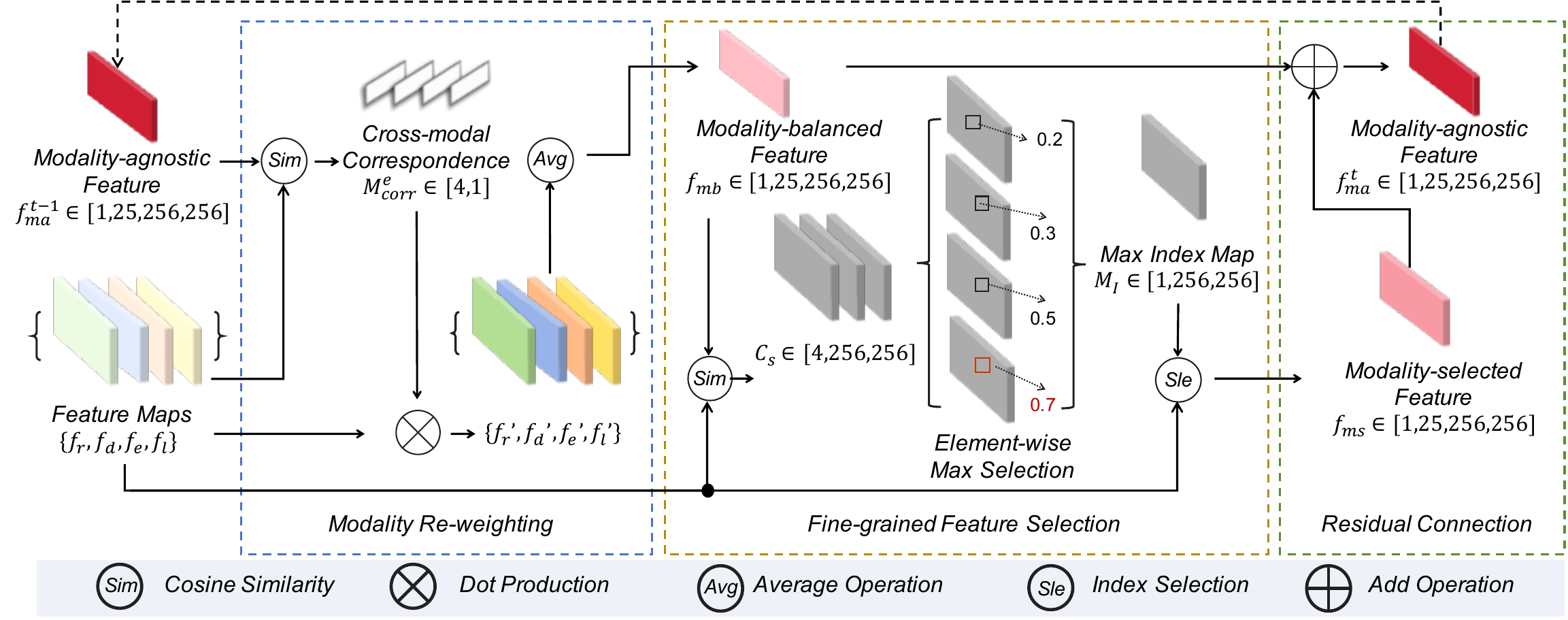}
    \caption{ We use the Modality-agnostic Feature Fusion Module to obtain modality-agnostic features for multi-modal segmentation.}
    \label{fig:MFFM}
\end{figure}
Building upon LSCD module, we then present the MFF module, a tailored architecture designed to \textbf{\textit{learn an optimal modality-agnostic representation across modalities}}, thus improving the robustness of multi-modal semantic segmentation in any visual conditions. 
The input of MFF module is the modality-agnostic feature $f_{ma}^{t-1}$ from the iteration at $t-1$, and the visual features \(\{f_r, f_d, f_e, f_l\} \). Note that the $f_{ma}^{0}$ is initialized by the average of multi-modal features \(\{f_r, f_d, f_e, f_l\}\).
As shown in Fig.~\ref{fig:MFFM}, we first compute the cross-modal correlation $M_{corr}^e$ by calculating the cosine similarity between the $f_{ma}^{t-1}$ and \(\{f_r, f_d, f_e, f_l\} \). Then \(M_{corr}^e\) is multiplied ($\otimes$) with visual features \(\{f_r, f_d, f_e, f_l\} \) to obtain the weighted feature maps. 
Then, the weighted feature maps are averaged to obtain the modality-balanced feature \(f_{mb}\).

Afterwards, MFF module computes the similarity between \(\{f_r, f_d, f_e, f_l\} \) and \(f_{mb}\) to obtain cross-modal similarity maps $C_S$. 
After that, we perform max similarity index selection for each element location across four similarity maps (denoted as gray maps in Fig.~\ref{fig:MFFM}) to get the max index map \(M_I\). 
MFF module uses the max index map \(M_I\) to select elements from \(\{f_r, f_d, f_e, f_l\} \) to obtain the modality-selected feature \(f_{ms}\). Finally, MFF module aggregates ($\oplus$) modality-balanced feature \(f_{mb}\) and modality-selected feature \(f_{ms}\) to get the modality-agnostic feature \(f_{ma}^t\) for iteration at $t$, which is subsequently fed into the segmentation head, resulting in the prediction \( P_{ma} \), from which the supervised loss is computed as:
\begin{equation}
\setlength{\abovedisplayskip}{3pt}
\setlength{\belowdisplayskip}{3pt}
\mathcal{L}_{sup} = \text{CE}(P_{ma}, Y),
\label{eq:l_sup}
\end{equation}
where \( \text{CE} \) represents the Cross-Entropy loss function. The modality-agnostic feature \(f_{ma}^t\) for iteration at $t$ is then used as input for iteration at $t+1$.
\subsection{Optimization, Training, and Validation}
The total training objective for learning the multi-modal semantic segmentation model containing three losses (Eq.\ref{eq:l_cr}, Eq.\ref{eq:l_se}, Eq.\ref{eq:l_sup}) is defined as:
\begin{equation}
\setlength{\abovedisplayskip}{3pt}
\setlength{\belowdisplayskip}{3pt}
    \mathcal{L} = \mathcal{L}_{se} + \mathcal{L}_{cr} + \mathcal{L}_{sup}. 
\end{equation}
Different from most multi-modal semantic segmentation frameworks~\cite{zhang2023delivering,zhang2022cmx} process multi-modal inputs with parallel modality-specific encoders, our method takes multi-modal data batch as input and simultaneously processes them with shared-weights encoders across modalities, the loss functions for each modality are integrated together before backward. During training, the multi-modal data flow is depicted in Fig.~\ref{overall}, and both the LSCD and MFF modules are included. For the validation phase with multi-modal data, Any2Seg performs solely the segmentation forward pass with MFF module. 

\noindent Multi-modal Semantic Segmentation (\textbf{MSS}) validation means that we pass all the modalities in training for validation.
\textbf{System-level} Modality-incomplete Semantic Segmentation (\textbf{MISS}) validation denotes that we evaluate all the possible combinations of input modalities and average all results to obtain final mean performance. \textbf{Sensor-level MISS} means that we evaluate the segmentation model with specific condition data from all modalities, such as cloudy weather.

\section{Experiments}
\begin{table}[t!]
\caption{Quantitative results in MSS setting across dual, triple, and quadruple modalities on DELIVER. R: RGB, D: Depth, E: Event, L: LiDAR.}
\renewcommand{\tabcolsep}{1pt}
\resizebox{\linewidth}{!}{
\begin{tabular}{c|c|c|c|c|c|c|c}
\midrule
Method & Modal & Backbone & mIoU & Method & Modal & Backbone & mIoU \\ \midrule
HRFuser~\cite{yuan2021hrformer} & \multirow{3}{*}{R} & HRFuser-T & 47.95 & HRFuser~\cite{yuan2021hrformer} & \multirow{5}{*}{R+L} & HRFuser-T & 43.13 \\ \cmidrule{1-1} \cmidrule{3-5} \cmidrule{7-8} 
SegFormer~\cite{xie2021segformer} &  & MiT-B0 & 52.10 & TokenFusion~\cite{wang2022multimodal} &  & \multirow{4}{*}{MiT-B2} & 53.01 \\ \cmidrule{1-1} \cmidrule{3-5} \cmidrule{8-8} 
SegFormer~\cite{xie2021segformer} &  & MiT-B2 & 57.20 & CMX~\cite{zhang2022cmx} &  &  & 56.37 \\ \cmidrule{1-5} \cmidrule{8-8} 
HRFuser~\cite{yuan2021hrformer} & \multirow{6}{*}{R+D} & HRFuser-T & 51.88 & CMNeXt~\cite{zhang2023delivering} &  &  & 58.04 \\ \cmidrule{1-1} \cmidrule{3-5} \cmidrule{8-8} 
FPT~\cite{liu2024fourier} &  & MultiMAE & 57.38 & Ours &  &  & 57.92  \\ \cmidrule{1-1} \cmidrule{3-8} 
TokenFusion~\cite{wang2022multimodal} &  & \multirow{4}{*}{MiT-B2} & 60.25 & HRFuser~\cite{yuan2021hrformer} & \multirow{3}{*}{R+D+E} & HRFuser-T & 51.83 \\ \cmidrule{1-1} \cmidrule{4-5} \cmidrule{7-8} 
CMX~\cite{zhang2022cmx} &  &  & 62.67 & CMNeXt~\cite{zhang2023delivering} &  & \multirow{2}{*}{MiT-B2} & 64.44 \\ \cmidrule{1-1} \cmidrule{4-5} \cmidrule{8-8} 
CMNeXt~\cite{zhang2023delivering} &  &  & 63.58 & \textbf{Ours} &  &  & \textbf{66.75} \textcolor{red}{+2.31} \\ \cmidrule{1-1} \cmidrule{4-8} 
\textbf{Ours} &  &  & \textbf{67.12} \textcolor{red}{+3.54} & HRFuser~\cite{yuan2021hrformer} & \multirow{3}{*}{R+D+L} & HRFuser-T & 52.72 \\ \cmidrule{1-5} \cmidrule{7-8} 
HRFuser~\cite{yuan2021hrformer} & \multirow{5}{*}{R+E} & HRFuser-T & 42.22 & CMNeXt~\cite{zhang2023delivering} &  & \multirow{2}{*}{MiT-B2} & 65.50 \\ \cmidrule{1-1} \cmidrule{3-5} \cmidrule{8-8} 
TokenFusion~\cite{wang2022multimodal} &  & \multirow{4}{*}{MiT-B2} & 45.63 & \textbf{Ours} &  &  & \textbf{67.20} \textcolor{red}{+1.70} \\ \cmidrule{1-1} \cmidrule{4-8} 
CMX~\cite{zhang2022cmx} &  &  & 56.62 & HRFuser~\cite{yuan2021hrformer} & \multirow{3}{*}{R+D+E+L} & HRFuser-T & 52.97 \\ \cmidrule{1-1} \cmidrule{4-5} \cmidrule{7-8} 
CMNeXt~\cite{zhang2023delivering} &  &  & 57.48 & CMNeXt~\cite{zhang2023delivering} &  & \multirow{2}{*}{MiT-B2} & 66.30 \\ \cmidrule{1-1} \cmidrule{4-5} \cmidrule{8-8} 
\textbf{Ours} &  &  & \textbf{57.83} \textcolor{red}{+0.35} & \textbf{Ours} &  &  & \textbf{68.25} \textcolor{red}{+1.95}\\ \midrule
\end{tabular}}
\label{tab:multimodal}
\end{table}

\noindent \textbf{Datasets.} 
DELIVER~\cite{zhang2023delivering} is a comprehensive multi-modal dataset featuring RGB, depth, LiDAR, and event data across 25 semantic categories, under various environmental conditions and sensor failures. We follow the official processing and split guidelines from ~\cite{zhang2023delivering}. 
MCubeS~\cite{liang2022multimodal} focuses on material segmentation, offering 20 categories with RGB, Near-Infrared (NIR), Degree of Linear Polarization (DoLP), and Angle of Linear Polarization (AoLP) images.
\begin{wraptable}{r}{0.5\textwidth}
\caption{MSS results on MCubeS~\cite{liang2022multimodal}.}
\renewcommand{\tabcolsep}{8pt}
\resizebox{0.5\textwidth}{!}{
\begin{tabular}{l|cc}
\toprule
Method & Modal & mIoU \\ \midrule
MCubeSNet~\cite{liang2022multimodal} & I & 33.70 \\ 
MCubeSNet~\cite{liang2022multimodal} & I-A & 39.10 \\
MCubeSNet~\cite{liang2022multimodal} & I-A-D & 42.00 \\
MCubeSNet~\cite{liang2022multimodal} & I-A-D & 42.86 \\ \midrule
CMNeXt~\cite{zhang2023delivering} (MiT-B2) & I-A & 48.82 \\
CMNeXt~\cite{zhang2023delivering} (MiT-B2) & I-A-D & 49.48 \\
CMNeXt~\cite{zhang2023delivering} (MiT-B2) & I-A-D-N & 51.54 \\ \midrule
Ours (MiT-B2) & I-A & \textbf{48.95} \\ 
Ours (MiT-B2) & I-A-D & \textbf{50.67} \\ 
Ours (MiT-B2) & I-A-D-N & \textbf{51.91} \\ 
\bottomrule 
\end{tabular}}
\label{tab:sotamcubes}
\end{wraptable}
\noindent \textbf{Implementation Details.} \\ Our Any2Seg is trained on 8 NVIDIA A800 GPUs, starting with a learning rate of 6 $e^{-5}$, adjusted by a poly strategy (power 0.9) across 200 epochs, including an initial 10-epoch warm-up at 0.1 times the learning rate. We use the AdamW optimizer with a batch size of 1 per GPU. Data augmentation includes random resizing (0.5-2.0 ratio), horizontal flipping, color jitter, gaussian blur, and cropping to 1024 $\times$ 1024 on ~\cite{zhang2023delivering} and 512 $\times$ 512 on ~\cite{liang2022multimodal}. 

\noindent \textbf{Multi-modal Semantic Segmentation (MSS)}
Tabs. \ref{tab:multimodal} and \ref{tab:sotamcubes} present a detailed quantitative evaluation, contrasting our proposed framework with previous SoTA methods. Any2Seg outperforms other methods in almost all multi-modal scenarios. Specifically, for dual-modality inputs on the DELIVER dataset, Any2Seg significantly exceeds the prior SoTA, CMNeXt \cite{zhang2023delivering}, with mIoU improvements of \textbf{+3.54} and \textbf{+0.35} for the R+D and R+E configurations, respectively. In scenarios using triplet modality inputs, our Any2Seg continues to lead, surpassing the SoTA with mIoU enhancements of \textbf{+2.31} and \textbf{+1.70} for the R+D+E and R+D+L modalities.
Most notably, by incorporating all four modalities, our Any2Seg not only maintains its leading edge but also reaches an unprecedented performance benchmark with a 68.25 mIoU, achieving a significant improvement of \textbf{+1.95} mIoU over the CMNeXt~\cite{zhang2023delivering}. Additionally, Any2Seg consistently delivers the highest performance on the MCubeS across various methods with only the proposed MFFM, as shown in Tab.~\ref{tab:sotamcubes}. 

\begin{table*}[t!]
\renewcommand{\tabcolsep}{1pt}
\caption{Results of system-level MISS evaluation. All methods are trained with four modalities, and the metric is mIoU for all numbers.}
\resizebox{\linewidth}{!}{
\begin{tabular}{c|ccccccccccccccc|c|c}
\toprule
\multirow{2}{*}{M.} & \multicolumn{15}{c|}{Modality-incomplete Validation on DELIVER~\cite{zhang2023delivering}} & \multirow{2}{*}{Mean} & \multirow{2}{*}{$\Delta$} \\ \cmidrule{2-16}
&  \multicolumn{1}{c}{R} & \multicolumn{1}{c}{D} & \multicolumn{1}{c}{E} & \multicolumn{1}{c}{L} & \multicolumn{1}{c}{RD} & \multicolumn{1}{c}{RE} & \multicolumn{1}{c}{RL} & \multicolumn{1}{c}{DE} & \multicolumn{1}{c}{DL} & \multicolumn{1}{c}{EL} & \multicolumn{1}{c}{\small{RDE}} & \multicolumn{1}{c}{\small{RDL}} & \multicolumn{1}{c}{\small{REL}} & \multicolumn{1}{c}{\small{DEL}} & \small{RDEL} &  \\ \midrule
\cite{zhang2023delivering} & \multicolumn{1}{c}{3.76} & \multicolumn{1}{c}{0.81} & \multicolumn{1}{c}{1.00} & \multicolumn{1}{c}{\textbf{0.72}} & \multicolumn{1}{c}{50.33} & \multicolumn{1}{c}{13.23} & \multicolumn{1}{c}{18.22} & \multicolumn{1}{c}{21.48} & \multicolumn{1}{c}{3.83} & \multicolumn{1}{c}{\textbf{2.86}} & \multicolumn{1}{c}{66.24} & \multicolumn{1}{c}{66.43} & \multicolumn{1}{c}{15.75} & \multicolumn{1}{c}{46.29} & 66.30 & 25.25 & -\\
Ours & \textbf{39.02} & \textbf{60.11} & \textbf{2.07} & 0.31 & \textbf{68.21} & \textbf{39.11} & \textbf{39.04} & \textbf{60.92} & \textbf{60.15} & 1.99 & \textbf{68.24} & \textbf{68.22} & \textbf{39.06} & \textbf{60.95} & \textbf{68.25} & \textbf{45.04} & \textbf{\textcolor{red}{+19.79}}\\ 
 \bottomrule
\end{tabular}}
\label{Tab:ArbitrarySeg_4modal}
\end{table*}
\begin{figure}[t!]
    \centering
    \includegraphics[width=\linewidth]{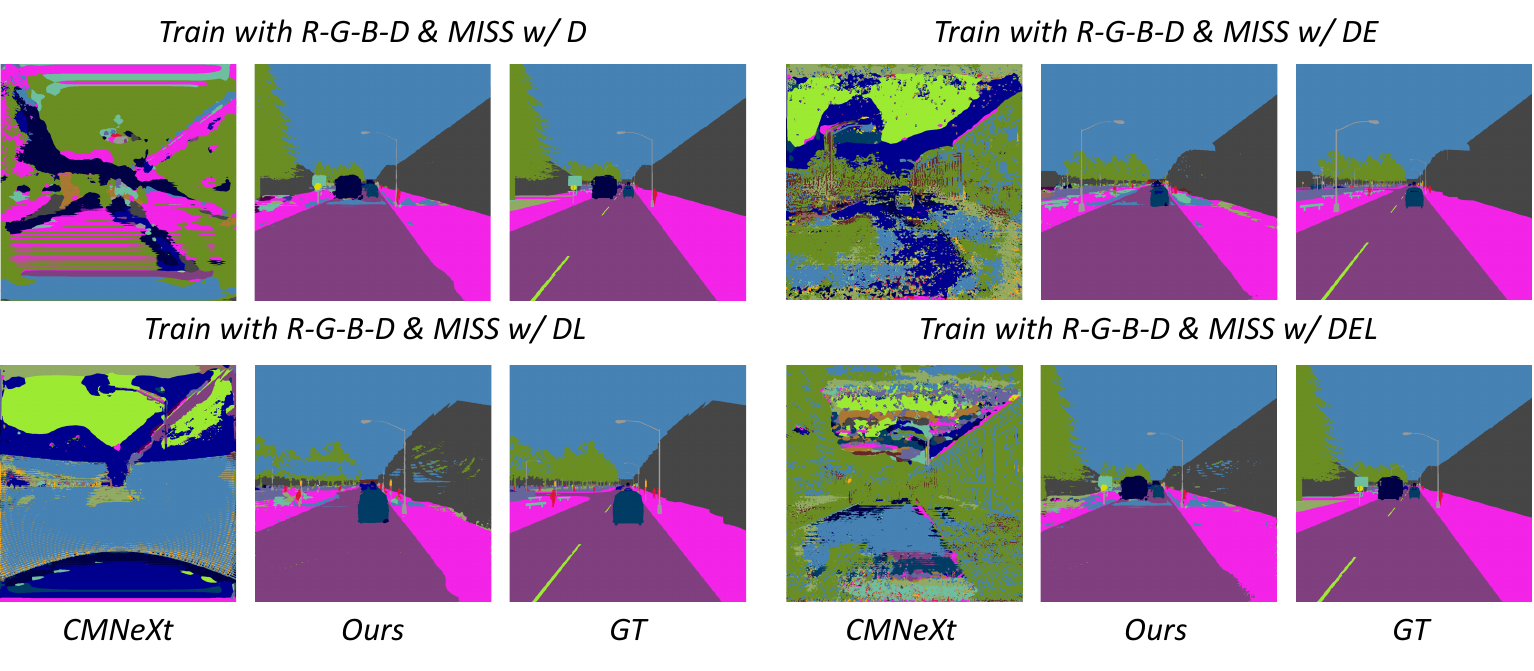}
    \caption{
    System-level Modality-Incomplete Semantic Segmentation (MISS) validation results on DEVLIER Dataset (D: depth; L: LiDAR; and E: Event).
    }
    \label{fig:modal_incom_vis}
\end{figure}

\noindent \textbf{Modality-Incomplete Semantic Segmentation (MISS)}\\
\noindent \textbf{System-level MISS:} 
As shown in Tab.~\ref{Tab:ArbitrarySeg_4modal}, Any2Seg significantly outperforms the CMNeXt framework~\cite{zhang2023delivering} in nearly all configurations, with a notable mean improvement of \textbf{+19.79} mIoU, especially in RGB and Depth combinations. 
While performance dips in scenarios with sparse modalities like Event and LiDAR, Any2Seg's design to harness multi-modal information enables it to excel in dense modality scenarios, significantly outdoing the benchmarks with a remarkable \textbf{60.15} mIoU in the DL (depth and LiDAR) modality compared to CMNeXt's 3.83 mIoU. This highlights the efficacy of Any2Seg in integrating multi-modal information to improve segmentation, as shown in Fig.~\ref{fig:modal_incom_vis}.

\noindent \textbf{Sensor-level MISS:} Our evaluation also encompasses sensor-level failures, leveraging the diverse adverse conditions present in the DELIVER dataset's training and validation sets, such as cloudiness and over-exposure. These conditions serve as effective data augmentation, against which Any2Seg consistently excels, surpassing prior state-of-the-art methods under challenging conditions, for instance, achieving 69.19 mIoU in cloudy weather, as depicted in Tab.~\ref{Tab:Adverse}. This performance attests to our LSCD and MFF modules' proficiency in minimizing modality gaps, fusing multi-modal data and extracting durable, modality-agnostic features for reliable segmentation across various conditions and sensor failure cases.

\begin{table*}[t!]
\renewcommand{\tabcolsep}{1pt}
\caption{\textbf{Snesor-level MISS results on DELIVER}. MB: Motion Blur; OE: Over-Exposure; UE: Under-Exposure; LJ: LiDAR-Jitter; and EL: Event Low-resolution. The \textbf{bold} and \underline{underline} denote the best and the second-best performance, respectively.}
\resizebox{\linewidth}{!}{
\begin{tabular}{llccccccccccccc}
\toprule
Method & Modality & Cloudy & Foggy & Night & Rainy & Sunny & MB & OE & UE & LJ & EL & Mean \\ \midrule
HRFuser~\cite{yuan2021hrformer} & \small{R}  & 49.26 & 48.64 & 42.57 & 50.61 & 50.47 & 48.33 & 35.13 & 26.86 & 49.06 & 49.88 & 47.95 \\
SegFormer~\cite{xie2021segformer} & \small{R} & 59.99 & 57.30 & 50.45 & 58.69 & 60.21 & 57.28 & 56.64 & 37.44 & 57.17 & 59.12 & 57.20 \\ \midrule
TokenFusion~\cite{wang2022multimodal} & \small{R-D} & 50.92 & 52.02 & 43.37 & 50.70 & 52.21 & 49.22 & 46.22 & 36.39 & 49.58 & 49.17 & 49.86 \\
CMX~\cite{zhang2022cmx} & \small{R-D} & 63.70 & 62.77 & 60.74 & 62.37 & 63.14 & 59.50 & 60.14 & 55.84 & 62.65 & 63.26 & 62.66 \\
HRFuser~\cite{yuan2021hrformer} & \small{R-D} & 54.80 & 51.48 & 49.51 & 51.55 & 52.12 & 50.92 & 41.51 & 44.00 & 54.10 & 52.52 & 51.88 \\ \midrule
HRFuser~\cite{yuan2021hrformer} & \small{R-D-E} & 54.04 & 50.83 & 50.88 & 51.13 & 52.61 & 49.32 & 41.75 & 47.89 & 54.65 & 52.33 & 51.83 \\
HRFuser~\cite{yuan2021hrformer} & \small{R-D-E-L} & 56.20 & 52.39 & 49.85 & 52.53 & 54.02 & 49.44 & 46.31 & 46.92 & 53.94 & 52.72 & 52.97 \\ \midrule
~\cite{zhang2023delivering} w/ S-2 & \small{R-D-E} & 60.02 & 56.16 & 54.03 & 54.82 & 58.29 & 53.70 & 57.04 & 51.98 & 58.54 & 55.87 & 56.05 \\
~\cite{zhang2023delivering} w/ S-2 & \small{R-D-L} & 68.28 & 63.28 & 62.64 & 63.01 & 66.06 & 62.58 & 64.44 & 58.73 & 65.37 & 65.80 & 64.02  \\
~\cite{zhang2023delivering} w/ S-2 & \small{R-D-E-L} & \underline{68.70} & 65.67 & 62.46 & \textbf{67.50} & 66.57 & 62.91 & 64.59 & 60.00 & 65.92 & 65.48 & 64.98 \\ \midrule
Ours w/ S-2 & \small{R-D-E} & 68.35 & 66.03 & 64.60 & 66.61 & 67.31 & \underline{63.62} & 64.61 & 62.34 & 65.55 & 64.20 & 65.32 \\
Ours w/ S-2 & \small{R-D-L} & 68.38 & \underline{67.18} & \underline{66.68} & 65.06 & \underline{67.73} & 63.34 & \textbf{66.11} & \underline{64.49} & \underline{67.54} & \underline{66.40} & \underline{66.29} \\ 
Ours w/ S-2 & \small{R-D-E-L} & \textbf{69.19} & \textbf{67.64} & \textbf{67.25} & \underline{67.00} & \textbf{68.25} & \textbf{66.23} & \underline{64.89} & \textbf{65.69} & \textbf{69.05} & \textbf{66.73} & \textbf{67.19} \\ 
\bottomrule
\end{tabular}}
\label{Tab:Adverse}
\end{table*}

\section{Ablation Study}

\subsection{Effectiveness of Loss Functions}
We first conduct ablation studies to evaluate the proposed losses. The results in Tab.~\ref{tab:lossab} show that all losses play positive roles.
For instance, adding $\mathcal{L}_{sup}$ to the baseline leads to gains of +1.04 mIoU, +1.16 F1, and +2.32 Acc. 
Obviously, using all three losses results in the highest performance, \ie, 63.74 mIoU, 73.33 F1, and 73.64 Acc, demonstrating their effectiveness.

\begin{table}[t!]
\renewcommand{\tabcolsep}{9pt}
\small
\caption{Ablation of different losses on DELIVER with SegFormer-B0 in MSS setting.}
\resizebox{\textwidth}{!}{
\begin{tabular}{cccc|cccccc}
\toprule
\multirow{2}{*}{$\mathcal{L}_{base}$} & \multirow{2}{*}{$\mathcal{L}_{se}$} & \multirow{2}{*}{$\mathcal{L}_{cr}$} & \multirow{2}{*}{$\mathcal{L}_{sup}$} & \multicolumn{6}{c}{RGB-Depth} \\ \cmidrule{5-10}
 &  &  &  & mIoU & $\Delta$ & F1 & $\Delta$ & Acc & $\Delta$ \\ \midrule
 \Checkmark  &  &  &  & 61.93 & - & 71.53 & - & 70.35 & -  \\  
 \Checkmark & \Checkmark &  &  & 62.89 & +0.96 & 72.52 & +0.99 & 72.14 & +1.79 \\ 
\Checkmark & & \Checkmark  &  & 62.86 & +0.93 & 72.36 & +0.83 & 72.81 & +2.46 \\ \midrule
& &  & \Checkmark & 62.97 & +1.04 & 72.69 & +1.16 & 72.67 & +2.32 \\ 
 & \Checkmark &  & \Checkmark & 63.41 & +1.48 & 73.15 & +1.62 & 73.19 & +2.84 \\  
& & \Checkmark & \Checkmark & 63.14 & +1.21 & 72.97 & +1.44 & 73.35 & +3.00 \\ 
 & \Checkmark & \Checkmark & \Checkmark & \textbf{63.74} & \textbf{+1.81} & \textbf{73.33} & \textbf{+1.80} & \textbf{73.64} & \textbf{+3.29} \\ \bottomrule
\end{tabular}}
\label{tab:lossab}
\end{table}

\subsection{Ablation of Components in MFF module}

To assess the effectiveness of MFFM's key components—feature re-weighting, fine-grained feature selection, and residual connections—we undertake a thorough evaluation across MSS and MISS settings within two datasets, covering four modalities as shown in Tab.~\ref{tab:MFFM}.
\begin{table*}[t!]
\caption{Ablation of MFF with SegFormer-B0 in both MSS and MISS settings.}
\renewcommand{\tabcolsep}{0.5pt}
\resizebox{\linewidth}{!}{
\begin{tabular}{c|c|c|cccccc|cccccc}
\toprule
\multicolumn{15}{c}{MCubeS Dataset with all four modalities} \\ \midrule
\multicolumn{1}{c}{Re-} & \multicolumn{2}{c}{Fusion} & \multicolumn{6}{c}{MSS} & \multicolumn{6}{c}{MISS} \\ \cmidrule{2-15}
 weight & Selection & Residual & mIoU & $\Delta$ & Acc & $\Delta$ & F1 & $\Delta$ & \multicolumn{1}{c}{mIoU} & $\Delta$ & \multicolumn{1}{c}{Acc} & $\Delta$ & \multicolumn{1}{c}{F1} & $\Delta$ \\ \midrule
\Checkmark &  & & 39.63 & - & 51.06 & - & 47.97 & - & 27.88 & - & 35.10 & - & 35.68 & - \\ \midrule
\Checkmark & \Checkmark &  & 42.49 & +2.86 & 54.90 & +3.84 & 52.09 & +4.12 & 29.10 & +1.22 & 38.24 & +3.14 & 36.56 & +0.88 \\ \midrule
 \Checkmark &  & \Checkmark & 42.86 & +3.23 & 54.91 & +3.85 & 52.62 & +4.65 & 25.10 & -2.78 & 33.25 & -1.85 & 32.95 & -2.73 \\ \midrule
\Checkmark & \Checkmark & \Checkmark & 44.20 & \textbf{+4.57} & 56.60 & \textbf{+5.54} & 54.12 & \textbf{+6.15} & 29.14 & \textbf{+1.26} & 38.81 & \textbf{+3.71} & 40.75 & \textbf{+5.07} \\ \midrule
\multicolumn{15}{c}{DELIVER Dataset with all four modalities} \\ \midrule
\multicolumn{1}{c}{Re-} & \multicolumn{2}{c}{Fusion} & \multicolumn{6}{c}{MSS} & \multicolumn{6}{c}{MISS} \\ \cmidrule{2-15}
 weight & Selection & Residual & mIoU & $\Delta$ & Acc & $\Delta$ & F1 & $\Delta$ & \multicolumn{1}{c}{mIoU} & $\Delta$ & \multicolumn{1}{c}{Acc} & $\Delta$ & \multicolumn{1}{c}{F1} & $\Delta$ \\ \midrule
\Checkmark &  & & 56.24 & - & 67.33 & - & 73.08 & - & 35.47 & - & 44.36 & - & 53.96 & -  \\ \midrule
\Checkmark & \Checkmark & & 61.61 & +5.37 & 70.98 & +3.65 & 71.61 & -1.47 & 44.49 & +9.02 & 52.79 & +8.43 & 54.28 & +0.32 \\ \midrule
\Checkmark & & \Checkmark & 60.96 & +4.72 & 71.35 & +4.02 & 74.12 & +1.04 & 36.22 & +0.75 & 45.06 & +0.70 & 52.51 & -1.45 \\ \midrule
\Checkmark & \Checkmark & \Checkmark & 62.97 & \textbf{+6.73} & 76.03 & \textbf{+8.70} & 72.67 & \textbf{-0.41} & 47.79 & \textbf{+12.32} & 57.12 & \textbf{+12.76} & 56.69 & \textbf{+2.79} \\ \bottomrule
\end{tabular}}
\label{tab:MFFM}
\end{table*}

\noindent \textbf{Fine-grained feature selection.}
Initially, modality-balanced re-weighting alone yields a mIoU of 39.63 and 56.24 in MSS validation on both datasets. The incorporation of our fine-grained feature selection further augments the performance significantly, enhancing the mIoU by +2.86 and +5.37, Acc by +3.84 and +3.65 on both datasets. Concurrently, MISS validation also exhibits improvements, with increases of +1.22 and +9.02 mIoU, +3.14 and +8.43 Acc. Obviously, the performance gains brought by our fine-grained feature selection are larger in modality-incomplete segmentation settings, which indicates that our proposed selection algorithm is efficient in selecting the fine-grained feature-agnostic features from the multi-modal inputs.
 
\noindent \textbf{Residual connections.}
The residual connections which aggregates the balanced feature with the selected feature, leads to significant gains of +3.23 and +4.72 in mIoU, +3.85 and +4.02 in Acc, and +4.65 and +1.04 in F1 during MSS on both datasets, as depicted in Tab.~\ref{tab:MFFM}. However, using the residual connection alone, without the fine-grained feature selection, results in mixed outcomes across all three segmentation metrics: -2.78 and +0.75 in mIoU, -1.85 and +0.70 in Acc, and -2.73 and -1.45 in F1. These findings highlight that while the residual connection mechanism enhances MSS, the fine-grained feature selection mechanism plays a pivotal role in improving MISS performance.

\begin{figure}[t!]
    \centering
    \includegraphics[width=\linewidth]{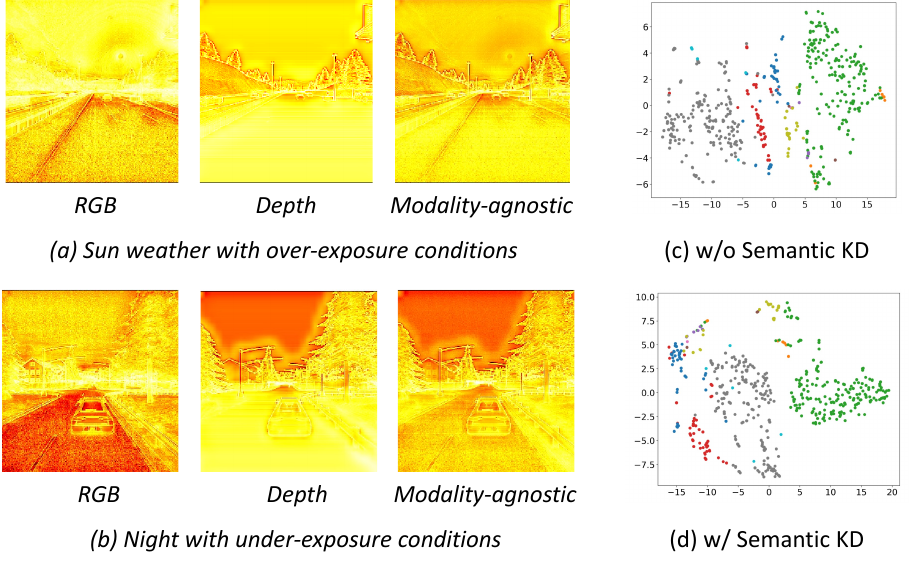}
    \caption{
    Visualization of multi-modal features under different conditions on DEVLIER. (a) sun with over-exposure, and (b) night with under-exposure. 
    }
    \label{fig:ab_MFFM_feat}
\end{figure}

\noindent \textbf{Effectiveness of the Overall MFF module.}
Integrating fine-grained feature selection with residual connections substantially enhances performance: in multi-modal validation, mIoU increases by +4.57 and +6.73 and Acc by +5.54 and +8.70; modality-incomplete segmentation sees mIoU gains of +1.26 and +12.32 and Acc of +3.71 and +12.76 across datasets. Meanwhile, using the selection mechanism alone results in a 61.61 mIoU in multi-modal segmentation on DELIVER, which rises to 62.97 with the residual connection, as shown in Tab.~\ref{tab:MFFM}. In modality-incomplete segmentation, combining all mechanisms increases performance from 36.22 to 47.79 mIoU. Meanwhile the qualitative results in Fig.~\ref{fig:ab_MFFM_feat} illustrates that modality-agnostic features merge the advantages of different modalities, \ie, RGB's textual information and depth cameras' lighting in-variance.
These results highlight MFF's crucial impact on extracting modality-agnostic features and improving segmentation accuracy in varied scenarios.

\subsection{Ablation of KD in Two Embedding Spaces}
\noindent \textbf{Effectiveness of Intra-modal Semantic KD.}
We conduct ablation studies on DELIVER, employing both single and dual modality inputs while contrasting the effects of solely using the basic loss $\mathcal{L}_{base}$ against the inclusion of our proposed semantic KD loss $\mathcal{L}_{se}$. The outcomes are shown in Tab.~\ref{tab:absemantic}, reveal that with a singular modality input, specifically RGB data, the integration of our semantic space KD results in notable improvements: a +1.36 increase in mAcc, a +1.64 enhancement in mIoU, and a +2.04 uplift in F1 score. In scenarios involving dual modalities, the application of our KD loss $\mathcal{L}_{se}$ yields further gains: +1.05 in mAcc, +0.81 in mIoU, and +0.66 in F1 score. 
Fig.~\ref{fig:ab_MFFM_feat} (c) and (d) illustrate that our semantic KD facilitates enhanced differentiation of semantics in the high-level space.
These findings robustly affirm the efficacy of implementing KD within the semantic embedding space.
\begin{table}[t!]
\centering
\begin{minipage}{0.49\linewidth}
\centering
\caption{Ablation study of $\mathcal{L}_{se}$.}
\renewcommand{\tabcolsep}{0.5pt}
\resizebox{\textwidth}{!}{
\begin{tabular}{ccccccccc}
\toprule
&\multirow{1}{*}{$\mathcal{L}_{base}$} & \multirow{1}{*}{$\mathcal{L}_{se}$} & \multirow{1}{*}{mAcc} & \multirow{1}{*}{$\Delta$} & \multirow{1}{*}{mIoU} & \multirow{1}{*}{$\Delta$} & \multirow{1}{*}{F1} & \multirow{1}{*}{$\Delta$}\\ \midrule
\multirow{2}{*}{R}&\Checkmark &  & 59.79 & - & 51.39 & - & 61.59 & - \\ 
&\Checkmark & \Checkmark & 61.15 & +\textbf{1.36} & 53.03 & +\textbf{1.64} & 63.63 & +\textbf{2.04}\\ 
\midrule
&\multirow{1}{*}{$\mathcal{L}_{base}$} & \multirow{1}{*}{$\mathcal{L}_{se}$} & \multirow{1}{*}{mAcc} & \multirow{1}{*}{$\Delta$} & \multirow{1}{*}{mIoU} & \multirow{1}{*}{$\Delta$} & \multirow{1}{*}{F1} & \multirow{1}{*}{$\Delta$}\\ \midrule
\multirow{2}{*}{R+E}&\Checkmark &  & 60.36 & - & 52.21 & - & 62.64 & - \\ 
&\Checkmark & \Checkmark & 60.20 & -0.16 & 53.11 & +\textbf{0.90} & 63.24 & +\textbf{0.60} \\ 
\bottomrule
\end{tabular}}
\label{tab:absemantic}
\end{minipage}
\hfill
\begin{minipage}{0.49\linewidth}
\centering
\caption{Ablation study of $\mathcal{L}_{cr}$.}
\resizebox{\textwidth}{!}{
\renewcommand{\tabcolsep}{0.5pt}
\begin{tabular}{ccccccccc}
\toprule
&\multirow{1}{*}{$\mathcal{L}_{base}$} & \multirow{1}{*}{$\mathcal{L}_{cr}$} & \multirow{1}{*}{mAcc} & \multirow{1}{*}{$\Delta$} & \multirow{1}{*}{mIoU} & \multirow{1}{*}{$\Delta$} & \multirow{1}{*}{F1} & \multirow{1}{*}{$\Delta$}\\ \midrule
\multirow{2}{*}{R+D}&\Checkmark &  & 61.93 & - & 71.53 & - & 70.35 & - \\ 
&\Checkmark & \Checkmark & 62.86 & +\textbf{0.93} & 72.36 & +\textbf{0.83} & 72.81 & +\textbf{2.46}\\ 
\midrule
&\multirow{1}{*}{$\mathcal{L}_{base}$} & \multirow{1}{*}{$\mathcal{L}_{cr}$} & \multirow{1}{*}{mAcc} & \multirow{1}{*}{$\Delta$} & \multirow{1}{*}{mIoU} & \multirow{1}{*}{$\Delta$} & \multirow{1}{*}{F1} & \multirow{1}{*}{$\Delta$}\\ \midrule
\multirow{2}{*}{R+E}&\Checkmark &  & 52.21 & - & 62.74 & - & 60.36 & - \\ 
&\Checkmark & \Checkmark & 53.26 & +\textbf{1.05} & 63.55 & +\textbf{0.81} & 61.02 & +\textbf{0.66} \\ 
\bottomrule
\end{tabular}}
\label{tab:losscr}
\end{minipage}
\end{table}

\noindent \textbf{Effectiveness of Inter-modal Correlation KD.}
We conduct ablation studies on DELIVER dataset with RGB-Event and RGB-Depth modalities while contrasting the effects of solely using the basic $\mathcal{L}_{base}$ against the inclusion of our proposed KD loss $\mathcal{L}_{cr}$, as shown in Tab.~\ref{tab:losscr}. Experimental results shows that in RGB-Depth setting, the integration of our inter-modal correlation KD leads to notable improvements: +0.93 mIoU, +0.83 F1, and +2.46 Acc score. Consistently, in RGB-Event setting, our $\mathcal{L}_{se}$ also achieves +0.90 mIoU improvement. 

\section{Conclusion}
In this paper, we made the first attempt to explore the potential of current MVLMs and introduced \textbf{Any2Seg} framework that can achieve robust segmentation from any combination of modalities in any visual conditions with the guidance from MVLMs. We proposed a Language-guided Semantic Correlation Distillation (\textbf{LSCD}) module for distilling semantic and cross-modal knowledge from MVLMs and a modality-agnostic feature fusion (\textbf{MFF}) module to transform multi-modal representation into modality-agnostic forms, ensuring robust segmentation. Our Any2Seg significantly outperformed previous multi-modal methods in both multi-modal and modality-incomplete settings with two public benchmarks, achieving new state-of-the-art segmentation performance.

\noindent \textbf{Limitation and Future Work.} Despite the contributions, there are still limitations of Any2Seg: (a) current MVLMs are pretrained at image-level rather than pixel-level to binding multi-modal data, restricting the dense pixel-level multi-modal segmentation performance; and (b) MSS and MISS performance remains sub-optimal in scenarios with sparse data inputs. Future research will concentrate on developing pixel-wise MVLMs and improving the modality-agnostic module to achieve uniform and enhanced performance across all scenarios.

\clearpage
\section{Acknowledgement}
This paper is supported by the National Natural Science Foundation of China (NSF) under Grant No. NSFC22FYT45, the Guangzhou City, University and Enterprise Joint Fund under Grant No.SL2022A03J01278, and Guangzhou Fundamental and Applied Basic Research (Grant Number: 2024A04J4072).

\bibliographystyle{splncs04}
\bibliography{main}
\end{document}


\title{Learning Modality-agnostic Representation for Semantic Segmentation from Any Modalities\\---Supplementary Material---}
\titlerunning{Abbreviated paper title}

\author{Xu Zheng\inst{1}\orcidlink{0000-0003-4008-8951} \and
Yuanhuiyi Lyu\inst{1}\orcidlink{0009-0004-1450-811X} 
\and
Lin Wang\inst{1,2}\orcidlink{0000-0002-7485-4493}\thanks{Corresponding author}}

\authorrunning{X. Zheng et al.}

\institute{Hong Kong University of Science and Technology, Guangzhou, China \\
\email{zhengxu128@gmail.com, yuanhuiyilv@hkust-gz.edu.cn}
\and
Hong Kong University of Science and Technology, Hong Kong, China \\
\email{linwang@ust.hk}\\
\url{https://vlislab22.github.io/Any2Seg/} 
}
\maketitle
\begin{abstract}
Due to the limited space in the main paper, this supplementary material provides an expansive elucidation of the proposed method and additional experimental results.
Sec.~\ref{sec:1} offers more details in implementing our AnySeg framework.
Sec.~\ref{sec:2} presents more experimental results, emphasizing both quantitative and qualitative assessments, and Sec.~\ref{sec:3} shows more ablation results.
\end{abstract}

\section{Implementation Details.} 
\label{sec:1}
\subsection{Datasets.} 
The \textbf{DELIVER} dataset~\cite{zhang2023delivering}, as introduced by Zhang \etal, represents a significant multi-modal segmentation dataset leveraging the CARLA simulator to include a variety of data types such as Depth, LiDAR, Views, Event, and RGB data. This dataset is notable for its dual-case offerings, which encompass diverse environmental conditions---cloudy, foggy, night, rainy, and sunny weather---and five specific scenarios of partial sensor failures. These environmental conditions pose perceptual challenges through variations in sunlight positioning and intensity, atmospheric diffusion, precipitation effects, and scene shading. The sensor failure scenarios comprehensively simulate malfunctions common to RGB cameras (Motion Blur, Over-Exposure, Under-Exposure), LiDAR jitter, and reduced resolution in event cameras, thereby providing a robust platform for testing and developing perception algorithms under a wide range of conditions.
Concurrently, \textbf{MCubeS}~\cite{liang2022multimodal} offers a multi-modal dataset designed for segmentation tasks across 20 categories, featuring pairs of RGB, Near-Infrared (NIR), Degree of Linear Polarization (DoLP), and Angle of Linear Polarization (AoLP) images. The dataset is divided into training, validation, and testing subsets, with 302, 96, and 102 image pairs respectively. This arrangement enables thorough evaluation and advancement of segmentation models, incorporating polarization data alongside traditional imaging modalities.

\subsection{Implementation Details.} Our Any2Seg is trained on 8 NVIDIA GPUs, starting with a learning rate of 6 $e^{-5}$, adjusted by a poly strategy (power 0.9) across 200 epochs, including an initial 10-epoch warm-up at 0.1 times the learning rate. We use the AdamW optimizer (epsilon 1$e^{-8}$, weight decay 1$e^{-2}$) with a batch size of 1 per GPU. Data augmentation includes random resizing (0.5-2.0 ratio), horizontal flipping, color jitter, gaussian blur, and cropping to 1024 $\times$ 1024 on ~\cite{zhang2023delivering} and 512 $\times$ 512 on ~\cite{liang2022multimodal}. 

\subsection{Metrics.}
In the evaluation of the performance of the proposed MAGIC framework, the assessment is anchored on three pivotal metrics: Intersection over Union (IoU), F1 score, and Accuracy (Acc), each providing unique insights into the model's segmentation capability.

\noindent \textbf{Intersection over Union (IoU)}
The IoU metric, also recognized as the Jaccard index, serves as a quantitative measure of the extent of overlap between the predicted and the ground truth segmentation maps. It is computed as the quotient of the intersection and the union of the predicted and ground truth segmentation areas. The IoU metric is normalized to range between 0 and 1, where a value closer to 1 denotes superior segmentation accuracy.

\noindent \textbf{F1 Score}
The F1 score, a harmonic mean of precision and recall, offers a balanced measure of the model's precision (the proportion of true positive results in all positive predictions) and recall (the proportion of true positive results among all actual positives). This metric is designed to provide a single measure to assess the precision-recall trade-off, with its value also ranging from 0 to 1, where higher values indicate more effective segmentation performance.

\noindent \textbf{Accuracy (Acc)}
Accuracy, expressed as a proportion, measures the fraction of pixels in the segmentation map that are correctly classified. This metric is calculated by dividing the tally of correctly classified pixels by the total pixel count within the map. Like IoU and F1 score, accuracy values span from 0 to 1, with higher values reflecting enhanced segmentation accuracy.

These metrics collectively facilitate a comprehensive evaluation of the MAGIC framework's performance, enabling a nuanced analysis of its segmentation efficacy across diverse conditions.

\section{Experimental Results}
\label{sec:2}

\begin{figure}[t!]
    \centering
    \includegraphics[width=\linewidth]{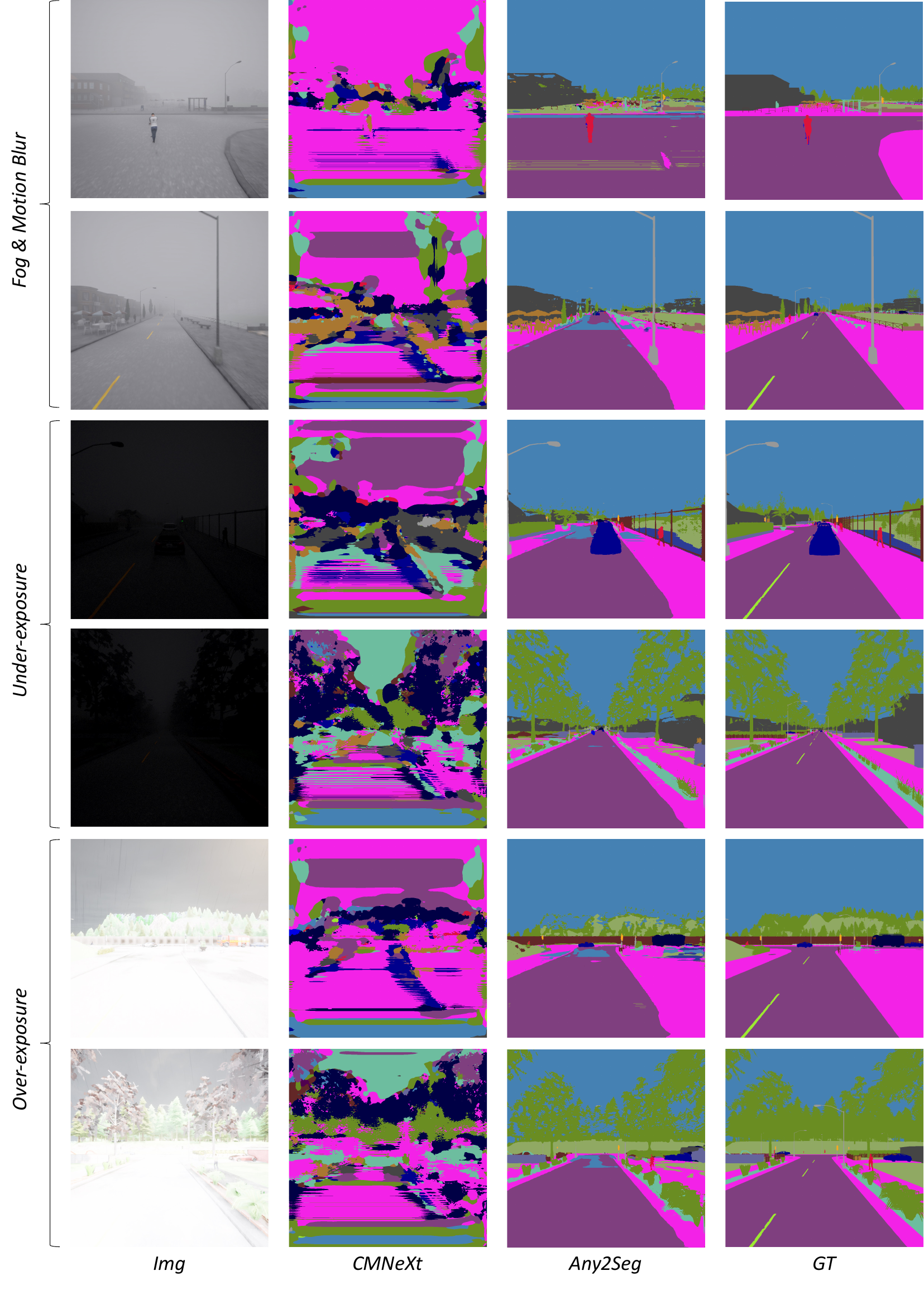}
    \caption{
    System-level Modality-Incomplete Semantic Segmentation (MISS) validation results on DEVLIER Dataset with only depth data input.
    }
    \label{fig:suppl_MISS_D}
\end{figure}

\begin{figure}[t!]
    \centering
    \includegraphics[width=\linewidth]{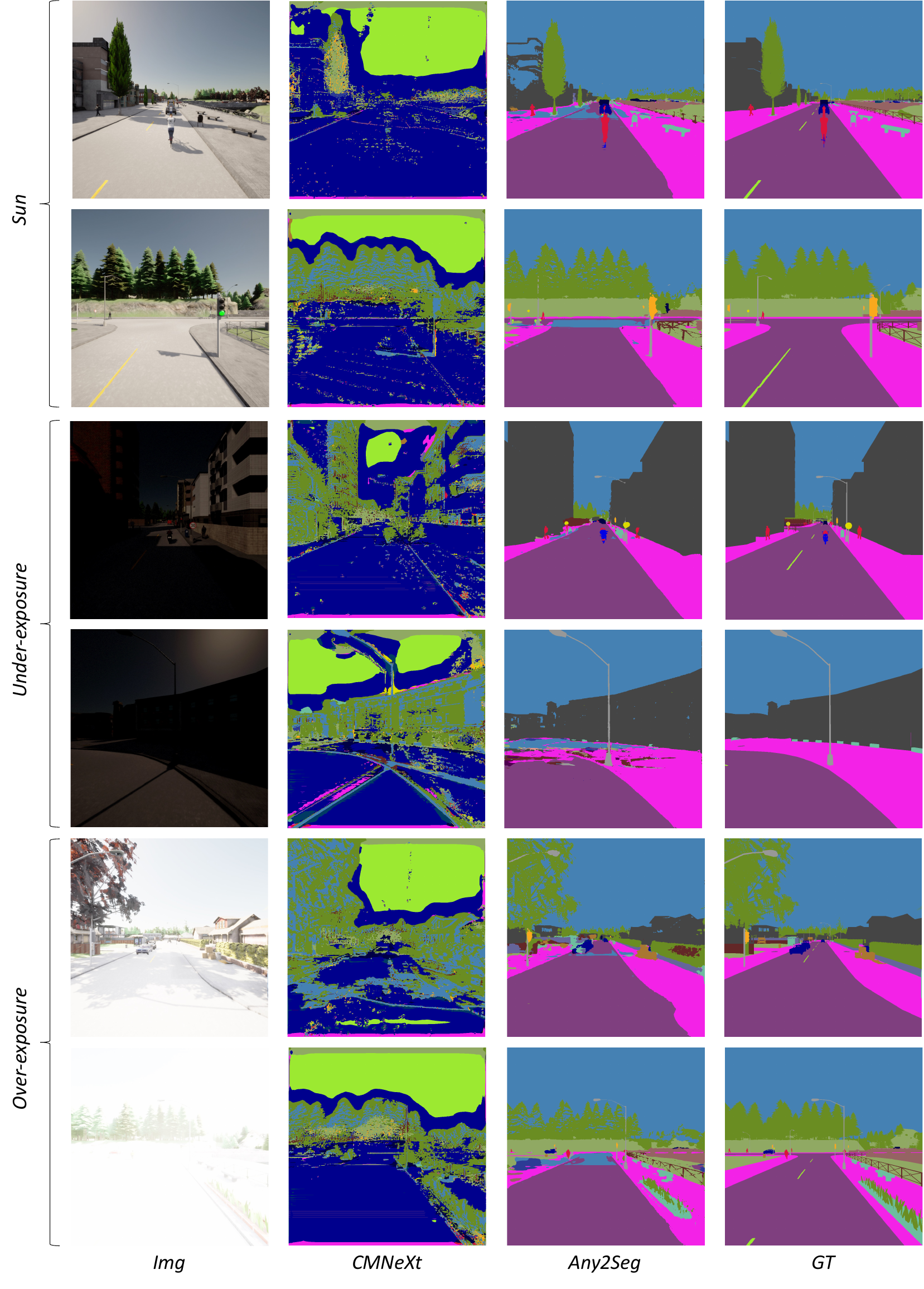}
    \caption{
    System-level Modality-Incomplete Semantic Segmentation (MISS) validation results on DEVLIER Dataset with depth and Event data input.
    }
    \label{fig:suppl_MISS_DE}
\end{figure}

\begin{figure}[t!]
    \centering
    \includegraphics[width=\linewidth]{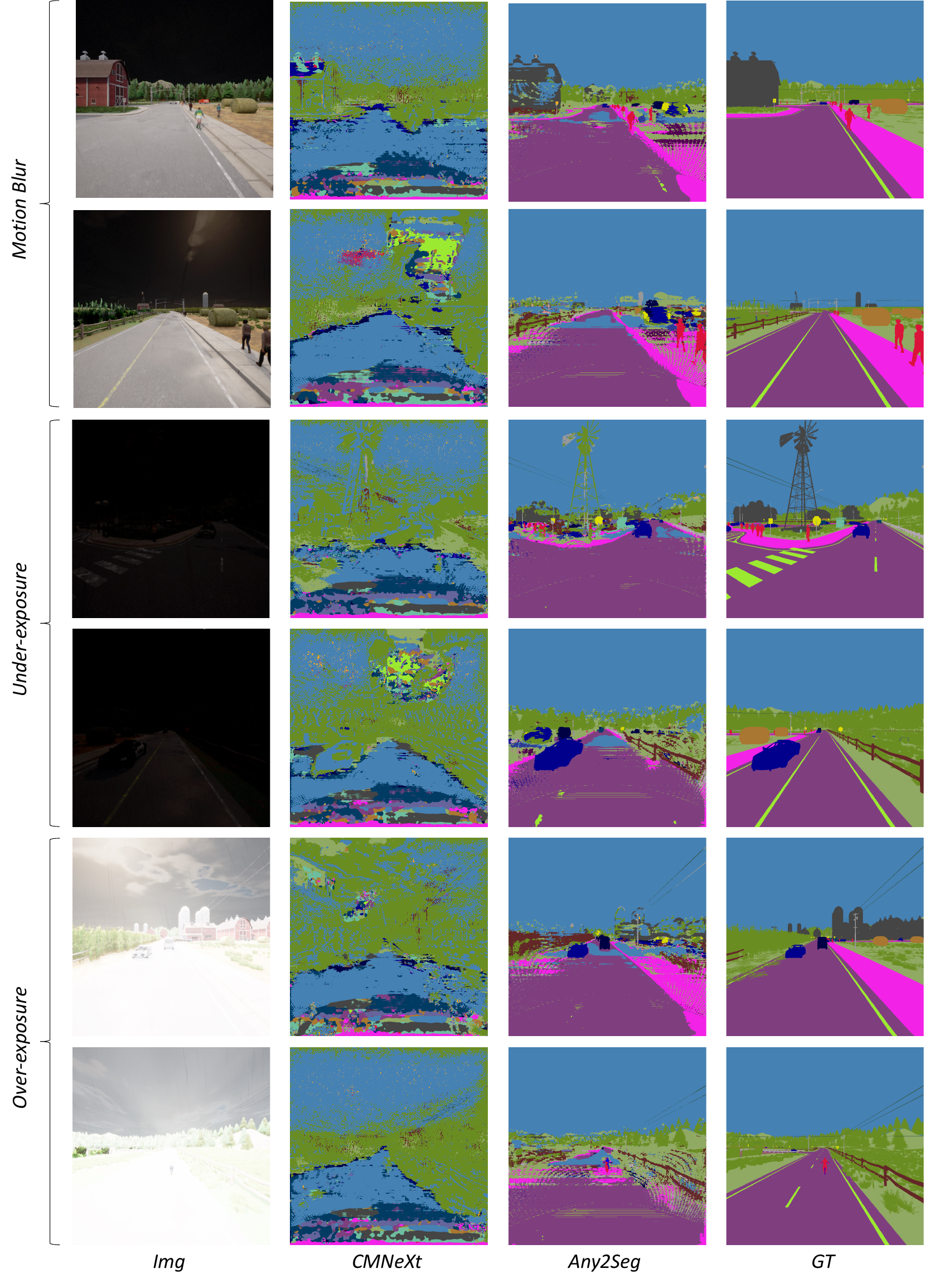}
    \caption{
    System-level Modality-Incomplete Semantic Segmentation (MISS) validation results on DEVLIER Dataset with depth, event, and LiDAR data input.
    }
    \label{fig:suppl_MISS_DEL}
\end{figure}

\begin{figure}[t!]
    \centering
    \includegraphics[width=\linewidth]{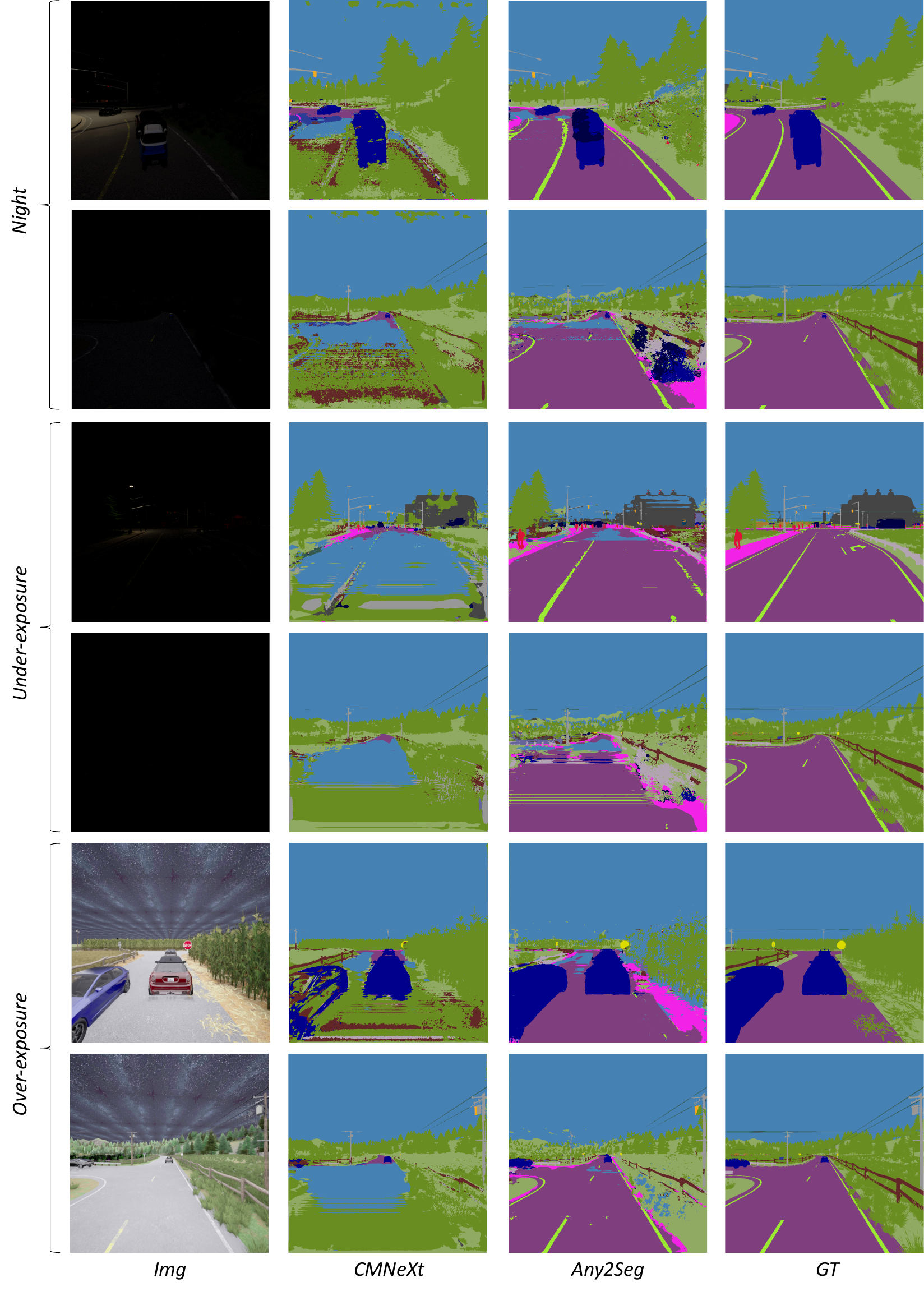}
    \caption{
    System-level Modality-Incomplete Semantic Segmentation (MISS) validation results on DEVLIER Dataset with RGR and depth data input.
    }
    \label{fig:suppl_MISS_RD}
\end{figure}

Tab.~\ref{Tab:DELIVERPerclass} shows the per-class results on DELIVER dataset, the training and validation is conducted with all four modalities.
Tab.~\ref{Tab:ArbitrarySeg_4modal} gives the qualitative results with three metrics in MISS validation.
Fig.~\ref{fig:suppl_MISS_D} presents a qualitative analysis comparing the performance of our method with the MISS validation criterion, employing solely depth data for inference. 
Fig.~\ref{fig:suppl_MISS_DE} depicts a qualitative comparison, utilizing the MISS validation, with the inference phase incorporating both depth and event data. 
Fig.~\ref{fig:suppl_MISS_DEL} offers a qualitative comparison, following the MISS validation framework, with inference leveraging depth, event, and LiDAR data. 
Fig.~\ref{fig:suppl_MISS_RD} provides a visualization comparison under the MISS validation scheme, utilizing RGB and depth data for inference. 

\begin{table*}[]
\caption{Per-class results on DELIVER dataset. The training and validation is conducted with four modalities: RGB, Depth, Event, and LiDAR. (Seg-B2: MiT-B2)}
\renewcommand{\tabcolsep}{0.5pt}
\resizebox{\linewidth}{!}{
\begin{tabular}{cccc|ccccccccccccc}
\toprule
 &  &  & Param & Build. & Fence & Other & Pede. & Pole & RL & Road & Side W. & Veget. & Cars & Wall & T. S. & Sky \\ \midrule
\multirow{8}{*}{IoU} & ~\cite{zhang2023delivering} & Seg-B2 & 58.73 & 89.41 & 43.12 & 0 & 76.51 & 75.13 & 85.91 & 98.18 & 82.27 & 88.97 & 84.98 & 69.39 & 70.57 & 99.43 \\ 
& Ours & Seg-B2 & 24.73 & 89.59 & 45.34 & 0 & 78.49 & 75.91 & 85.87 & 98.33 & 84.51 & 88.95 & 91.56 & 58.30 & 71.97 & 99.45 \\ 
& & & $\Delta$ & \textbf{+0.18} & \textbf{+2.22} & 0 & \textbf{+1.98} & \textbf{+0.78} & -0.04 & -0.15 & \textbf{+2.24} & -0.02 & \textbf{+6.58} & -11.09 & \textbf{+1.40} & \textbf{+0.02} \\ \cmidrule{2-17}
& Method & Backbone & Param & Ground & Bridge & Rail T. & G. R. & Traffic L. & Static & Dynamic & Water & Terr. & Two W. & Bus & Truck & Mean \\ \cmidrule{2-17}
 & ~\cite{zhang2023delivering} & Seg-B2 & 58.73 & 1.31 & 53.61 & 61.48 & 55.01 & 84.22 & 33.58 & 32.30 & 23.96 & 83.94 & 77.33 & 92.25 & 94.55 & 66.30 \\ 
& Ours & Seg-B2 & 24.73 & 2.30 & 59.80 & 66.59 & 63.50 & 85.04 & 37.36 & 33.41 & 46.22 & 82.65 & 78.08 & 91.61 & 91.30 & 68.25 \\ 
& & & $\Delta$ & \textbf{+0.99} & \textbf{+6.19} & \textbf{+5.11} & \textbf{+8.49} & \textbf{+0.82} & \textbf{+3.78} & \textbf{+1.11} & \textbf{+22.26} & -1.29 & \textbf{+0.75} & -0.64 & -3.25 & \textbf{+1.95} \\ \midrule
Metric & Method & Backbone & Param & Build. & Fence & Other & Pede. & Pole & RL & Road & Side W. & Veget. & Cars & Wall & T. S. & Sky \\ \midrule
\multirow{8}{*}{F1} & ~\cite{zhang2023delivering} & Seg-B2 & 58.73 & 94.41 & 60.26 & 0 & 86.69 & 85.80 & 92.42 & 99.08 & 90.28 & 94.16 & 91.88 & 81.93 & 82.74 & 99.71 \\ 
& Ours & Seg-B2 & 24.73 & 94.51 & 62.39 & 0 & 87.95 & 86.30 & 92.40 & 99.16 & 91.60 & 94.15 & 95.60 & 73.66 & 83.70 & 99.72 \\ 
& & & $\Delta$ & \textbf{+0.10} & \textbf{+2.13} & 0 & \textbf{+1.26} & \textbf{+0.50} & -0.02 & \textbf{+0.08} & \textbf{+1.32} & -0.01 & \textbf{+3.72} & -8.27 & \textbf{+0.96} & \textbf{+0.01} \\  \cmidrule{2-17} 
 & Method & Backbone & Param & Build. & Fence & Other & Pede. & Pole & RL & Road & Side W. & Veget. & Cars & Wall & T. S. & Sky \\ \cmidrule{2-17} 
& ~\cite{zhang2023delivering} & Seg-B2 & 58.73 & 2.59 & 69.80 & 76.14 & 70.98 & 91.43 & 50.28 & 48.83 & 38.66 & 91.27 & 87.22 & 95.97 & 97.20 & 75.19 \\
& Ours & Seg-B2 & 24.73 & 4.51 & 74.85 & 79.95 & 77.67 & 91.92 & 54.39 & 50.08 & 63.22 & 90.50 & 87.69 & 95.62 & 95.45 & 77.08 \\ 
& & & $\Delta$ & \textbf{+1.92} & \textbf{+5.05} & \textbf{+3.81} & \textbf{+6.69} & \textbf{+0.49} & \textbf{+4.11} & \textbf{+1.25} & \textbf{+24.56} & -0.77 & \textbf{+0.47} & -0.35 & -1.75 & \textbf{+1.89} \\  
\midrule
Metric & Method & Backbone & Param & Build. & Fence & Other & Pede. & Pole & RL & Road & Side W. & Veget. & Cars & Wall & T. S. & Sky \\ \midrule
\multirow{8}{*}{Acc} & ~\cite{zhang2023delivering} & Seg-B2 & 58.73 & 98.24 & 57.18 & 0 & 87.58 & 85.25 & 89.70 & 98.95 & 95.36 & 94.19 & 98.65 & 87.98 & 83.33 & 99.75 \\ 
 & Ours & Seg-B2 & 24.73 & 98.45 & 61.19 & 0 & 89.77 & 85.52 & 90.05 & 98.95 & 95.35 & 93.94 & 97.32 & 78.65 & 83.21 & 99.75\\ 
& & & $\Delta$ & \textbf{+0.21} & \textbf{+4.01} & 0 & \textbf{+2.19} & \textbf{+0.27} & \textbf{+0.35} & 0 & -0.01 & -0.25 & -1.33 & -9.33 & -0.12 & 0\\  \cmidrule{2-17}
& Method & Backbone & Param & Ground & Bridge & Rail T. & G. R. & Traffic L. & Static & Dynamic & Water & Terr. & Two W. & Bus & Truck & Mean \\ \cmidrule{2-17}
 & ~\cite{zhang2023delivering} & Seg-B2 & 58.73 & 2.00 & 61.91 & 75.28 & 56.60 & 88.71 & 35.32 & 50.35 & 24.05 & 93.65 & 86.86 & 96.13 & 97.12 & 73.77 \\ 
& Our & Seg-B2 & 24.73 & 6.08 & 63.18 & 73.55 & 65.63 & 89.22 & 40.67 & 53.38 & 52.08 & 93.36 & 85.79 & 95.22 & 98.73 & 75.56 \\ 
 & & & $\Delta$ & \textbf{+4.08} & \textbf{+1.27} & -1.73 & \textbf{+9.03} & \textbf{+0.51} & \textbf{+5.35} & \textbf{+3.03} & \textbf{+28.03} & -0.29 & -1.07 & -0.91 & \textbf{+1.61} & \textbf{+1.79} \\  
 \bottomrule
\end{tabular}}
\label{Tab:DELIVERPerclass}
\end{table*}

\begin{table*}[t!]
\renewcommand{\tabcolsep}{1pt}
\caption{Results of system-level MISS evaluation. All methods are trained with four modalities, and the metric is mIoU for all numbers.}
\resizebox{\linewidth}{!}{
\begin{tabular}{c|ccccccccccccccc|c|c}
\toprule
\multirow{2}{*}{M.} & \multicolumn{15}{c|}{Modality-incomplete Validation on DELIVER~\cite{zhang2023delivering} (\textbf{mIoU})} & \multirow{2}{*}{Mean} & \multirow{2}{*}{$\Delta$} \\ \cmidrule{2-16}
&  \multicolumn{1}{c}{R} & \multicolumn{1}{c}{D} & \multicolumn{1}{c}{E} & \multicolumn{1}{c}{L} & \multicolumn{1}{c}{RD} & \multicolumn{1}{c}{RE} & \multicolumn{1}{c}{RL} & \multicolumn{1}{c}{DE} & \multicolumn{1}{c}{DL} & \multicolumn{1}{c}{EL} & \multicolumn{1}{c}{\small{RDE}} & \multicolumn{1}{c}{\small{RDL}} & \multicolumn{1}{c}{\small{REL}} & \multicolumn{1}{c}{\small{DEL}} & \small{RDEL} &  \\ \midrule
\cite{zhang2023delivering} & \multicolumn{1}{c}{3.76} & \multicolumn{1}{c}{0.81} & \multicolumn{1}{c}{1.00} & \multicolumn{1}{c}{\textbf{0.72}} & \multicolumn{1}{c}{50.33} & \multicolumn{1}{c}{13.23} & \multicolumn{1}{c}{18.22} & \multicolumn{1}{c}{21.48} & \multicolumn{1}{c}{3.83} & \multicolumn{1}{c}{\textbf{2.86}} & \multicolumn{1}{c}{66.24} & \multicolumn{1}{c}{66.43} & \multicolumn{1}{c}{15.75} & \multicolumn{1}{c}{46.29} & 66.30 & 25.25 & -\\
Ours & \textbf{39.02} & \textbf{60.11} & \textbf{2.07} & 0.31 & \textbf{68.21} & \textbf{39.11} & \textbf{39.04} & \textbf{60.92} & \textbf{60.15} & 1.99 & \textbf{68.24} & \textbf{68.22} & \textbf{39.06} & \textbf{60.95} & \textbf{68.25} & \textbf{45.04} & \textbf{\textcolor{red}{+19.79}}\\ \midrule
\multirow{2}{*}{M.} & \multicolumn{15}{c|}{Modality-incomplete Validation on DELIVER~\cite{zhang2023delivering} (\textbf{Acc})} & \multirow{2}{*}{Mean} & \multirow{2}{*}{$\Delta$} \\ \cmidrule{2-16}
&  \multicolumn{1}{c}{R} & \multicolumn{1}{c}{D} & \multicolumn{1}{c}{E} & \multicolumn{1}{c}{L} & \multicolumn{1}{c}{RD} & \multicolumn{1}{c}{RE} & \multicolumn{1}{c}{RL} & \multicolumn{1}{c}{DE} & \multicolumn{1}{c}{DL} & \multicolumn{1}{c}{EL} & \multicolumn{1}{c}{\small{RDE}} & \multicolumn{1}{c}{\small{RDL}} & \multicolumn{1}{c}{\small{REL}} & \multicolumn{1}{c}{\small{DEL}} & \small{RDEL} &  \\ \midrule
\cite{zhang2023delivering} & 53.43 & 67.66 & 5.12 & \textbf{4.07} & 72.98 & 53.17 & 52.84 & 67.65 & 67.58 & 4.47 & 72.93 & 72.90 & 52.53 & 67.54 & 72.84 & 52.51 & -\\
Ours & \textbf{57.82} & \textbf{69.06} & \textbf{7.06} & 3.90 & \textbf{75.57} & \textbf{57.84} & \textbf{57.66} & \textbf{69.90} & \textbf{69.05} & \textbf{7.06} & \textbf{75.59} & \textbf{75.54} & \textbf{57.67} & \textbf{69.87} & \textbf{75.56} & \textbf{55.28} & \textbf{\textcolor{red}{+2.77}}\\ \midrule
\multirow{2}{*}{M.} & \multicolumn{15}{c|}{Modality-incomplete Validation on DELIVER~\cite{zhang2023delivering} (\textbf{F1})} & \multirow{2}{*}{Mean} & \multirow{2}{*}{$\Delta$} \\ \cmidrule{2-16}
&  \multicolumn{1}{c}{R} & \multicolumn{1}{c}{D} & \multicolumn{1}{c}{E} & \multicolumn{1}{c}{L} & \multicolumn{1}{c}{RD} & \multicolumn{1}{c}{RE} & \multicolumn{1}{c}{RL} & \multicolumn{1}{c}{DE} & \multicolumn{1}{c}{DL} & \multicolumn{1}{c}{EL} & \multicolumn{1}{c}{\small{RDE}} & \multicolumn{1}{c}{\small{RDL}} & \multicolumn{1}{c}{\small{REL}} & \multicolumn{1}{c}{\small{DEL}} & \small{RDEL} &  \\ \midrule
\cite{zhang2023delivering} & 45.02 & 67.36 & 3.11 & \textbf{2.37} & 74.36 & 45.55 & 45.75 & 67.47 & 67.38 & 2.65 & 74.37 & 74.36 & 46.02 & 67.44 & 74.37 & 50.51 & -\\
Ours & \textbf{50.35} & \textbf{69.98} & \textbf{3.80} & 0.60 & \textbf{77.06} & \textbf{50.58} & \textbf{50.50} & \textbf{71.26} & \textbf{70.01} & \textbf{3.66} & \textbf{77.08} & \textbf{77.06} & \textbf{50.67} & \textbf{71.28} & \textbf{77.08} & \textbf{53.40} & \textbf{\textcolor{red}{+2.89}}\\
 \bottomrule
\end{tabular}}
\label{Tab:ArbitrarySeg_4modal}
\end{table*}

\section{Ablation Study}
\label{sec:3}

\begin{figure}[t!]
    \centering
    \includegraphics[width=0.85\linewidth]{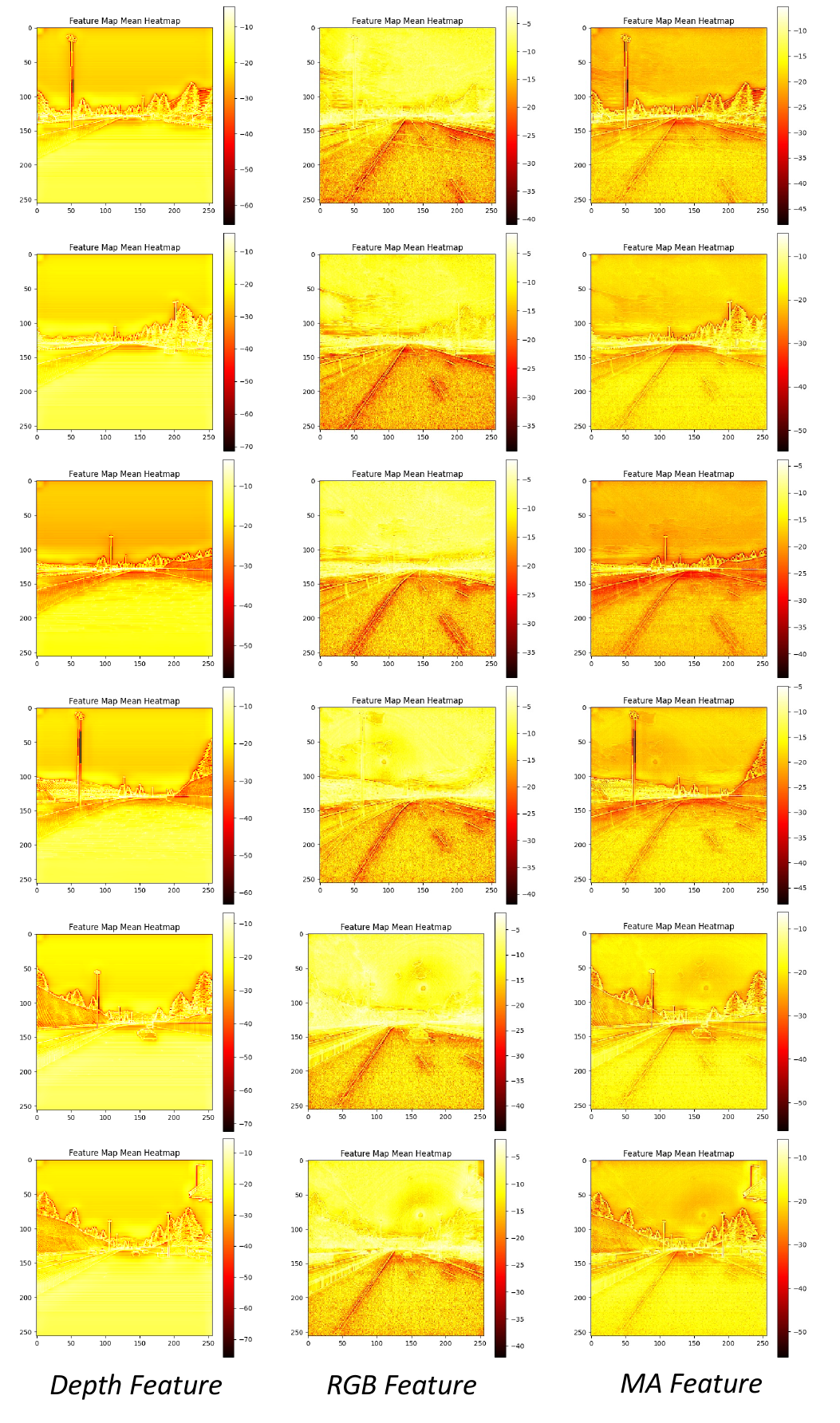}
    \caption{
    Visualization of multi-modal features under different conditions on DEVLIER. MA feature: Modality-agnostic feature.
    }
    \label{fig:suppl_feat_compare}
\end{figure}

\begin{figure}[t!]
    \centering
    \includegraphics[width=\linewidth]{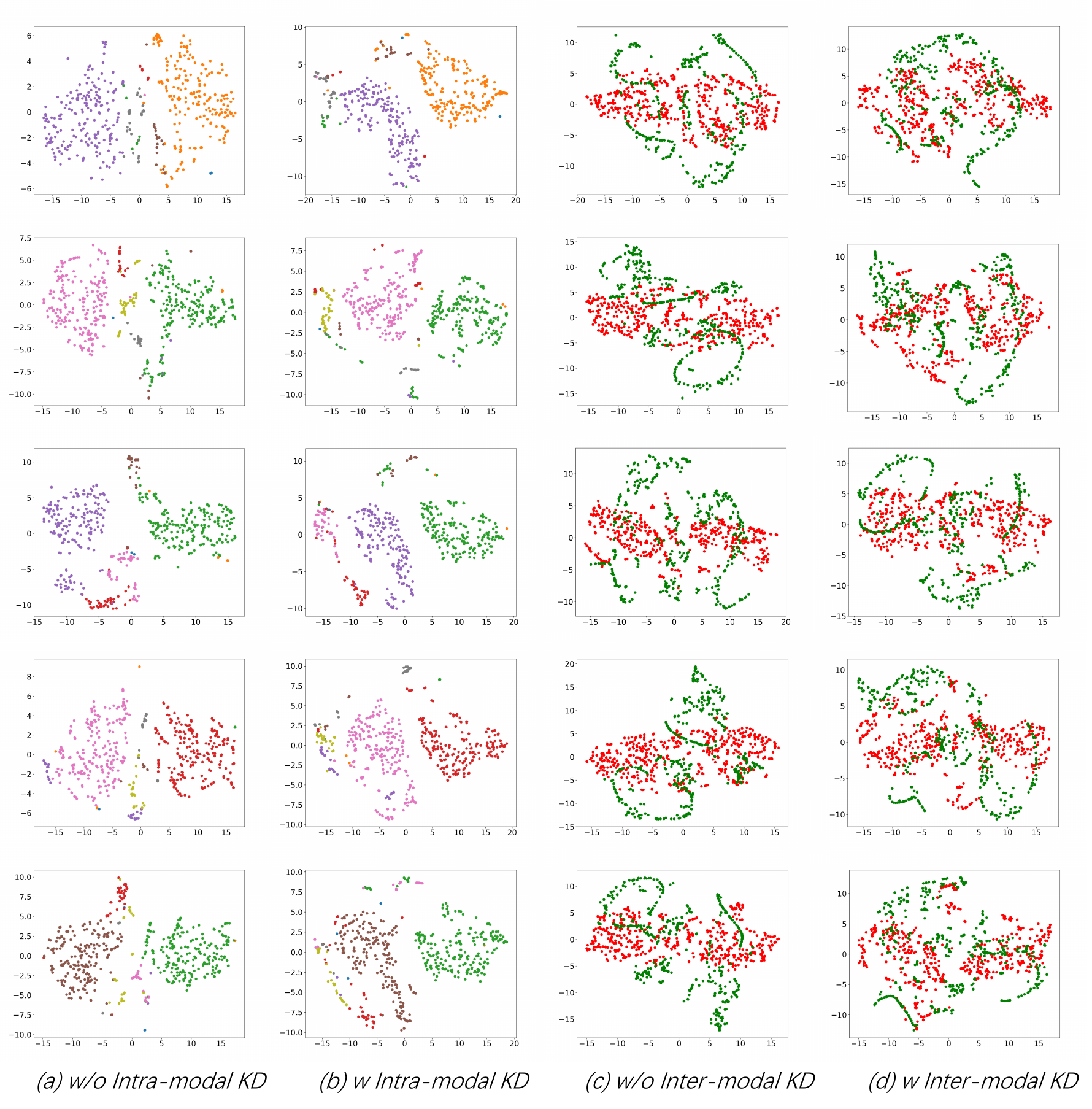}
    \caption{
    \textbf{t-SNE visualization}. Red and green points in (c) and (d) represent RGB and Depth features, respectively. In (a) and (b), colors indicate distinct semantic classes.
    }
    \label{fig:suppl_tsne}
\end{figure}

Fig.~\ref{fig:suppl_feat_compare} shows more visualization of RGB, depth, and our obtained modality-agnostic features.
Fig.~\ref{fig:suppl_tsne} presents more t-SNE visualization to ablate the effectiveness of our proposed inter- and intra-modal knowledge distillation.

\clearpage
%
\bibliographystyle{splncs04}
\bibliography{main}